%% file: main.tex
\pdfoutput=1
\documentclass[11pt]{article}
\usepackage[preprint]{acl}
\usepackage{times}
\usepackage{latexsym}
\usepackage[T1]{fontenc}
\usepackage[utf8]{inputenc}
\usepackage{microtype}
\usepackage{inconsolata}

\usepackage{microtype}
\usepackage{graphicx}
\usepackage{subfigure}
\usepackage{booktabs} 

\usepackage{hyperref}

\usepackage{amsmath}
\usepackage{amssymb}
\usepackage{mathtools}
\usepackage{amsthm}
\usepackage{enumitem}
\usepackage{amsfonts}  
\usepackage{fancyhdr}  
\usepackage{setspace}  
\usepackage{multirow}
\usepackage{graphicx}
\usepackage{caption}
\usepackage{booktabs}
\usepackage{siunitx}
\usepackage{tcolorbox}
\usepackage{listings}

\newcommand{\bbR}{\mathbb{R}}

\definecolor{codegreen}{rgb}{0,0.6,0}
\definecolor{codegray}{rgb}{0.5,0.5,0.5}
\definecolor{codepurple}{rgb}{0.58,0,0.82}
\definecolor{backcolour}{rgb}{0.95,0.95,0.92}
\lstdefinestyle{mystyle}{
    backgroundcolor=\color{backcolour},   
    commentstyle=\color{codegreen},
    keywordstyle=\color{magenta},
    numberstyle=\tiny\color{codegray},
    stringstyle=\color{codepurple},
    basicstyle=\ttfamily\footnotesize,
    breakatwhitespace=false,         
    breaklines=true,                 
    captionpos=b,                    
    keepspaces=true,                 
    showspaces=false,                
    showstringspaces=false,
    showtabs=false,                  
    tabsize=2
}

\usepackage{algorithm}
\usepackage{algorithmic}

\usepackage{stfloats}
\usepackage{makecell}  
\usepackage{enumerate} 
\usepackage{color}     
\usepackage{titlesec}
\usepackage{array} 
\newcolumntype{C}{>{\centering\arraybackslash}p{1.5cm}} 
\usepackage{threeparttable}  
\usepackage{booktabs}         

\usepackage[capitalize,noabbrev]{cleveref}

\theoremstyle{plain}

\theoremstyle{definition}

\theoremstyle{remark}

\usepackage[textsize=tiny]{todonotes}

\title{GraphThought: Graph Combinatorial Optimization with Thought Generation}

\newcounter{template}
\newcounter{algorithm_saved}
\makeatletter
\newenvironment{template}[1][htb]{%
	\renewcommand{\ALG@name}{Program}
	\setcounter{algorithm_saved}{\value{algorithm}} 
	\setcounter{algorithm}{\value{template}}
	\begin{algorithm}[#1]%
	}{\end{algorithm}
	\setcounter{template}{\value{algorithm}}
	\setcounter{algorithm}{\value{algorithm_saved}}
}
\makeatother

\author{
    {\bf Zixiao Huang\textsuperscript{1}}, 
    {\bf Lifeng Guo\textsuperscript{1,5}}, 
    {\bf Wenhao Li\textsuperscript{2}}, 
    {\bf Junjie Sheng\textsuperscript{3}}, 
    {\bf Chuyun Shen\textsuperscript{1}},\\
    {\bf Haosheng Chen\textsuperscript{1}}, 
    {\bf Xiangfeng Wang\textsuperscript{1,5}}, 
    {\bf Changhong Lu\textsuperscript{1,5}}, 
    {\bf Bo Jin\textsuperscript{2,4}}\thanks{
        Emails: \{zxhuang, cyshen, 51265901016, jarvis\}@stu.ecnu.edu.cn; 
        \{lfguo, chlu\}@math.ecnu.edu.cn; 
        \{whli, bjin\}@tongji.edu.cn; 
        xfwang@sei.ecnu.edu.cn
    } \\\\
    \textsuperscript{1}East China Normal University \quad
    \textsuperscript{2}Tongji University \quad
    \textsuperscript{3}Independent Researcher \\
    \textsuperscript{4}Shanghai Research Institute for Intelligent Autonomous Systems, Tongji University \\
    \textsuperscript{5}Key Laboratory of Mathematics and Engineering Applications, MoE, East China Normal University \\
}

\begin{document}

\maketitle



\begin{abstract}
    
Graph combinatorial optimization (GCO) problems are central to domains like logistics and bioinformatics. 
While traditional solvers dominate, large language models (LLMs) offer new possibilities for structured reasoning, yet struggle with complex GCO tasks requiring rigorous combinatorial analysis and multi-step deduction, often producing hallucinated steps. 
We first formalize the \textit{Optimal Thoughts Design} (OTD) problem, which provides a structured guidance for producing high-quality intermediate reasoning steps.  
Building on this formulation, we introduce \textit{GraphThought}, a novel framework that generates effective reasoning sequences through either heuristic-guided \textit{forward} search or solver-aligned \textit{backward} reasoning. 
By fine-tuning LLMs on these structured thought sequences\footnote{Dataset is available at \url{https://anonymous.4open.science/r/GraphThought-7CFE}.}, we develop Llama-GT, an 8B-parameter model that achieves state-of-the-art performance on the GraphArena benchmark, outperforming significantly larger models like DeepSeek-V3.
Our results demonstrate that when scaffolded with structured reasoning priors, principled thought generation can significantly enhance LLM performance on GCO tasks without requiring increased model scale.

\end{abstract}

\section{Introduction}\label{sec:intro}
Graph combinatorial optimization (GCO)problems constitute a broad class of computationally challenging tasks defined over graph structures, ranging from shortest‐path and optimal tour routing to selecting maximum independent sets (MIS) or minimum vertex covers (MVC). 
These problems, which have driven fundamental advances in discrete mathematics and computer science for decades \citep{Kuhn1955, Kruskal1956, Ford1956}, manifest across diverse application domains including logistics optimization, electronic circuit design, bioinformatics, and social network analysis \citep{tang2025grapharena}.
Many canonical GCO tasks—most notably the traveling salesman problem (TSP) and MIS—remain NP‐hard, motivating ongoing research into both exact and approximation techniques.

\begin{figure}[ht]
\begin{center}
\centerline{\includegraphics[width=\columnwidth]{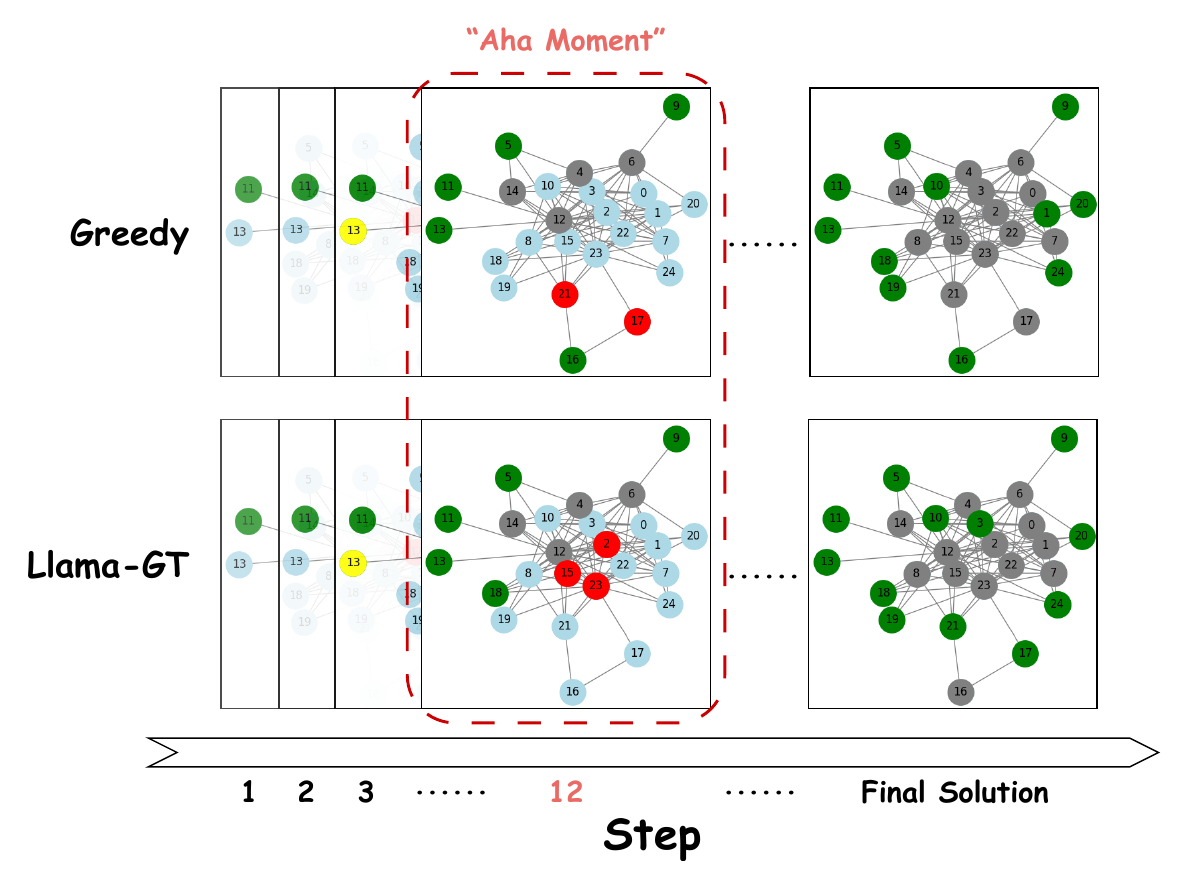}}
\caption{Comparative Analysis of Greedy Algorithm versus LLM-Generated Solution for MIS. Upper: Stepwise greedy selection process (e.g., selecting node 16, then removing neighbors 21 and 17), yielding an \textbf{11}-node solution. Lower: Llama-GT-generated solution employing adaptive heuristics (e.g., prioritizing node 18 and removing neighbors 2, 15, 23), achieving a superior \textbf{12}-node independent set. \textit{Color legend:} yellow = isolated nodes; green = selected nodes; red = neighbors of current selection; gray = removed nodes.}
\label{fig:greedy_case}
\end{center}
\vspace{-1.25cm}
\end{figure}

Traditional approaches to GCO have predominantly relied on human-designed algorithms and heuristics. 
Over decades, researchers have developed various methodologies including greedy algorithms, local search techniques, branch-and-bound approaches, and approximation schemes, which demonstrate notable effectiveness on moderate-scale instances \citep{Kuhn1955, Ford1956, Kruskal1956, paschos2014applications}. 
While such manually engineered approaches often produce satisfactory solutions, they require substantial manual effort and domain-specific expertise. 
Furthermore, heuristics constrained by fixed patterns (e.g., selecting minimum-degree nodes in MIS) often fail to account for global optimality, potentially leading to suboptimal outcomes through locally optimal choices, which can be observed in Figure~\ref{fig:greedy_case}.

In recent years, deep neural network–based approaches have attracted considerable attention for addressing GCO problems by learning solutions in a data-driven manner. 
Methods based on deep reinforcement learning and graph neural networks have demonstrated the ability to learn heuristics for specific tasks like TSP and MIS, often treating the problem solver as a black-box model optimized end-to-end \citep{Jin2024, Jiang2024, Liu2023, Iklassov2024, Lehnert2024}. 
However, a major drawback is that each problem typically requires a specialized network architecture or input encoding – for instance, a model architecture tailored for routing problems may not directly work for MIS, necessitating bespoke representation engineering per task \citep{Jiang2024, Jin2024}. 
Moreover, pure learning-based solvers often fail to exploit obvious problem-specific priors that humans would incorporate. 

Recent work has explored using LLMs for GCO through natural language reasoning, where models process graph descriptions and generate stepwise solutions by leveraging embedded algorithmic knowledge \citep{Wang2024a, Ren2024, Li2024a, Liu2023a, Luo2024}. 
This paradigm offers flexibility across problems through textual problem phrasing and recall of training-exposed strategies \citep{Li2024a, Ren2024}, achieving promising zero-shot/few-shot performance on small-scale tasks \citep{Wang2024a, Liu2023a}. 
However, LLMs significantly trail dedicated solvers on NP-hard problems like MIS and graph coloring, with benchmarks revealing poor solution optimality and validity even for advanced models like GPT-4 \citep{tang2025grapharena, Wang2024a}. 
Critical challenges include hallucinated reasoning steps \citep{Liu2023a, Luo2024} and unreliable multi-step deduction, despite mitigation attempts through structured prompts \citep{Luo2024, Ouyang2024, Skianis2024, Sanford2024, Li2024b}. 
While various approaches employ prompt engineering, code generation, or non-text embeddings \citep{jin2024large, chen2024exploring, liu2023towards, ren2024survey}, their performance remains substantially inferior to commercial solvers like Gurobi, highlighting limitations in directly mapping problems to solutions without addressing GCO's inherent long-horizon reasoning requirements.
This raises an intuitive question: 
\textit{Can we incorporate the search principles of classical GCO solvers into the LLM's output process?}

    Related ideas have been proposed to enhance the general problem-solving capabilities of LLMs, such as simulating human cognitive processes by generating intermediate thoughts prior to final responses. 
    Methods like chain-~\citep{wei2022chain}, tree-~\citep{yao2024tree} and graph-of-thoughts~\citep{besta2024graph} prompting encourage step-by-step reasoning~\citep{besta2024demystifying}.
    While these techniques are often effective, they can sometimes degrade performance due to self-enforcing~\citep{huang2024large}. 
    Moreover, techniques that work well on one dataset may not generalize to others due to variations in the type of reasoning involved (e.g., spatial reasoning vs. mathematical reasoning)~\citep{searchformer,wu2024thinking}.
    Similar limitations appear in recent GCO-focused works~\citep{luo2024graphinstruct, searchformer, chen2024graphwiz, zhang2024can, ouyang2024gundam, gandhi2024stream}, which construct thoughts by unrolling search algorithms or generating them via LLMs, followed by supervised fine-tuning.
    However, thought generation in GCO presents a unique challenge: 
    \textit{For NP-hard or NP-complete problems, no efficient traditional (heuristic) search methods exist, rendering forward thought construction infeasible.}

\begin{figure*}[htb]
\begin{center}
    \includegraphics[width=0.9\textwidth]{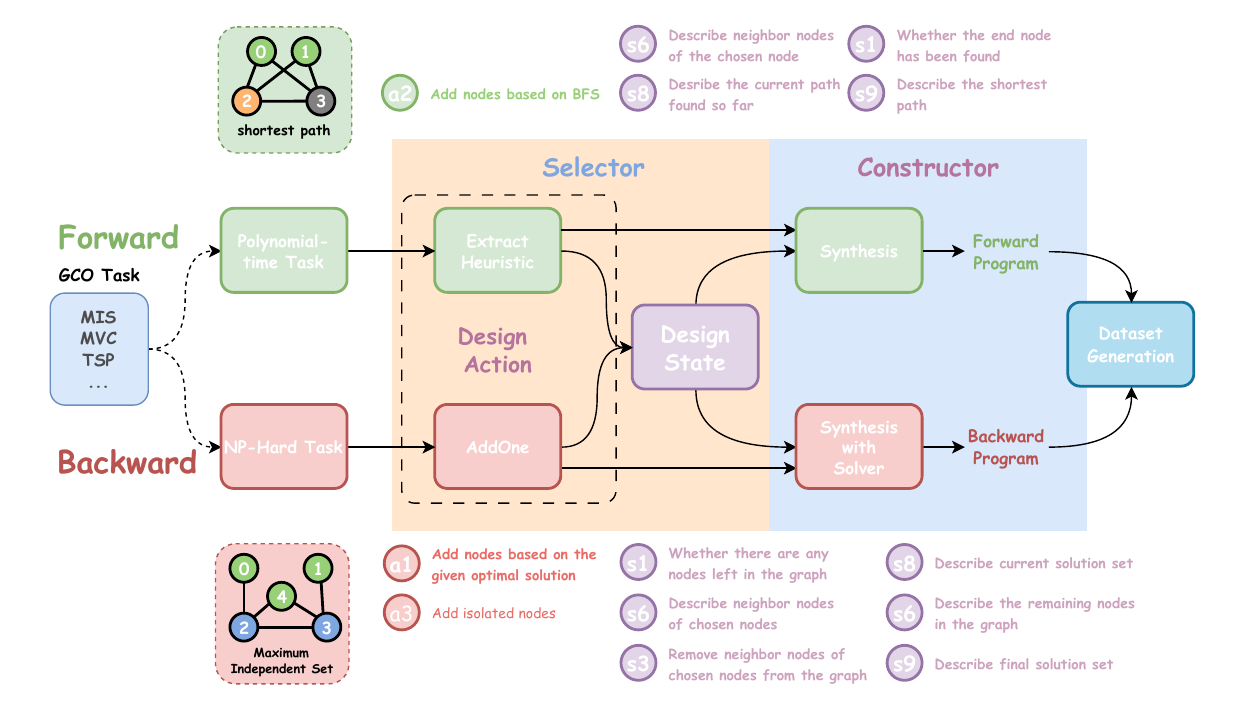}
    \caption{The \textit{Forward} and \textit{Backward} MTP frameworks for constructing the \textbf{Selector} and \textbf{Constructor} modules in GraphThought through distinct pathways. The \textbf{Selector} extracts action-state pairs from spaces $\mathbb{A}$ and $\mathbb{S}$, while the \textbf{Constructor} synthesizes an executable program. Two paradigm examples are shown: (1) Forward MTP for polynomial problems (e.g., shortest path) uses heuristic extraction of action thoughts (e.g., BFS-based node addition) to guide state selection; (2) Backward MTP for NP-hard problems (e.g., MIS) reconstructs reasoning steps through solver-derived solutions and reverse analysis. Both frameworks generate programs via systematic composition of thought sets, with backward MTP requiring combinatorial solvers. Appendix~\ref{app:thought_case} shows detailed reasoning steps, with full thought spaces described in Appendix~\ref{app:a_s}.}
\label{fig:framework}
\end{center}
\vspace{-0.75cm}
\end{figure*}

To address this, the Optimal Thoughts Design (OTD) problem is formally defined through a unified thought representation framework that systematically encodes search principles into LLM reasoning processes. The OTD formulation introduces action thought spaces $\mathbb{A}$ and state thought spaces $\mathbb{S}$ to model reasoning trajectories, enabling the generation of high-quality solution paths for GCO tasks. Building on this foundation, GraphThought—a novel framework that consists of forward and backward thought generation paradigms—is proposed. The forward mode employs heuristic-guided mechanisms for tractable problems, while the backward mode implements solver-guided backtracking techniques to handle NP-hard challenges.

Through extensive experiments, Llama-GT is developed by fine-tuning the Llama-3-8B-Instruct model using reasoning data generated by the proposed framework. 
Comprehensive evaluations across diverse GCO tasks demonstrate remarkable performance improvements, encompassing both polynomial-time solvable and NP-hard problems. 
Testing was conducted on graphs varying from small-scale (few nodes) to medium-scale (dozens of nodes). 
Llama-GT achieves solution quality that approaches the optimality bounds of commercial solvers like Gurobi in certain tasks, while retaining the flexibility of LLM-based reasoning.


The main contributions are as follows: 1) We formalize the OTD problem with state thought space $\mathbb{S}$ and action thought space $\mathbb{A}$, which facilitates the systematic generation of reasoning thoughts. 2) Within the GraphThought framework, we propose dual thought generation frameworks: a forward (heuristic-guided) one and a backward (solver-guided) one for GCO problems. 3) The state-of-the-art performance of our approach is evidenced by the fine-tuning of Llama-3-8B-Instruct, which achieves superior accuracy on GraphArena, outperforming both proprietary and open-source LLMs by significant margins. 
and approaching the performance of the commercial solver Gurobi.

\section{Problem Formulation}


This section formalizes the core problems addressed in this work. Despite the rapid progress of large language models (LLMs) across scientific and engineering domains, a key challenge remains in adapting them to complex mathematical tasks requiring advanced reasoning and domain-specific knowledge.

Combinatorial optimization (CO) problems form a class of mathematical optimization problems where the objective is to identify the optimal solution from a finite or countably infinite set of discrete candidates. 
Solving CO problems typically requires carefully crafted heuristic algorithms, often derived through rigorous mathematical analysis. 
These algorithms integrate sophisticated strategies to balance solution quality and computational efficiency.
These algorithms embody sophisticated strategies to balance solution quality and computational efficiency. 
A fundamental challenge lies in incorporating the search principles of heuristic algorithms into learnable reasoning thoughts for LLMs, enabling knowledge transfer while preserving algorithmic efficiency. 
We focus on GCO problems because graphs naturally model a variety of CO problems. 
A graph represents discrete entities through nodes and their connections. For a graph $G = (V,E)$, the node set $V = \{v_1, \ldots, v_n\}$ contains $n$ nodes, while the edge set $E = \{e_1,\ldots, e_m\}$ defines $m$ adjacency relationships. An edge $e=(v_i,v_j)$ denotes the adjacency of nodes $v_i$ and $v_j$.
Formally, let ${\cal{F}}: ({\rm {LLM}}, D) \rightarrow {\rm {LLM}}_D$ denote the supervised fine-tuning process mapping a foundation LLM to its fine-tuned version ${\rm {LLM}}_D$ through the training dataset $D$. The evaluation metric ${\cal{M}}: {\rm {LLM}}_D \rightarrow \bbR^+$ measures model performance.
The core optimization challenge is formalized as:
\begin{equation}\label{prob:core_problem}
    D^* = \arg \min_{D} {\cal{M}} \big( {\cal{F}} (\text{LLM},D) \big),
\end{equation} where $D^*$ is the optimal training dataset given LLM, ${\cal{F}}$, and ${\cal{M}}$. 
The formulation \eqref{prob:core_problem} constitutes a CO challenge over the exponentially large space of possible training datasets $D$. Three key difficulties emerge: 1) The discrete solution space prohibits gradient-based optimization; 2) The black-box nature of ${\cal{M}}$ prevents analytical evaluation; 3) The computational cost of evaluating ${\cal{M}}$ grows super-linearly with instance size.
It becomes important to efficiently compute high-quality approximate optimal solutions of \eqref{prob:core_problem}.

Thoughts of solving specific instances of math problems can be viewed as high-quality training data for fine-tuning LLMs and thus developing enhanced thoughts is critical for the creation of superior training datasets.
In the following, we will propose a series of thought generation methods to establish thoughts as one kind of approximate suboptimal solutions for the formulation \eqref{prob:core_problem}.
 
The representation and design of thoughts are typically a promising research direction in order to enhance the reasoning capabilities of LLMs.
The thought generation process can be modeled as to solve an approximate optimal solution $D' \subset \hat{D}$ for the formulation \eqref{prob:core_problem}, where $\hat{D}$ denotes the set of all training data with thoughts as content. For GCO problems, there are mainly two kinds of thoughts. One kind is to decide the following available action executed on the instance and the other kind is to show the current solving state.
A hierarchical decision problem in generating these kinds of thoughts is govern as: 
\begin{equation}\label{prob:otd}
    \left\{\begin{array}{l}
        {\hbox{Choose optimal}} \ A^*\subset \mathbb{A} , S^* \subset \mathbb{S}, \\
        {\hbox{Constructing program}} \  {\cal{P}}\ {\hbox{by}}\ A^*, S^*.
    \end{array}\right.
\end{equation}
where $\mathbb{A}$ and $\mathbb{S}$ denote the action thought space and state thought space for a GCO problem, respectively. 
The first item is to choose optimal $A^*$ from $\mathbb{A}$ and optimal $S^*$ from $\mathbb{S}$ according to the characteristics of the GCO problem, where $A$ encapsulates core algorithmic operations while $S$ maintains dynamic problem-solving states. The second item is to incorporate all chosen thoughts of $A^*$ and $S^*$ into a program template $\cal{P}$, which is used to generate specific thoughts in solving a GCO instance. 

In summary, the OTD problem is to generate the optimal thought set $A^*$ and $S^*$ to construct a program $\cal{P}$ for a GCO problem, which is the key problem to be addressed in this work. 
To emphasize, the optimal thoughts dataset in OTD might not be the optimal solution of the formulation \eqref{prob:core_problem}, but it can be considered as a high-quality approximate solution.

\section{The GraphThought Framework}\label{sec:method}

\subsection{The Overall Framework}
\label{subsec:framework}

GraphThought is introduced as a novel framework for fine-tuning LLMs through reasoning thought generation, as formalized in Algorithm~\ref{algo:g_t_d}. The architecture of the framework comprises two core modules that collaboratively generate task-specific reasoning programs by selecting state and action thoughts and constructing programs, denoted as $\textbf{Selector}$ and $\textbf{Constructor}$, respectively.

\noindent - {\bf{Selector}}: For a given task $\tau$, this module performs dynamic selection of \textit{action-state pairs} $(A, S)$ from predefined action and state thought space $\mathbb{A}$ and $\mathbb{S}$. The selection mechanism adaptively adjusts its strategies to capture the essential characteristics of $\tau$, ensuring context-aware component selection;

\noindent - {\bf{Constructor}}: This module operates as a program synthesis engine that methodically assembles the selected $(A, S)$ into an executable program ${\cal{P}}$. 

For the predefined action thought space $\mathbb{A}$ and state thought space $\mathbb{S}$, we systematically derive nine state representations and sixteen action operators through rigorous analysis of task-solving processes. These elements constitute the fundamental components of our framework, where $\mathbb{A}$ primarily represents operations on the solution set (e.g., adding nodes or edges), while $\mathbb{S}$ captures both graph operations (e.g., node insertion/removal) and state descriptions (e.g., remaining nodes/edges, current solution status). Complete specifications are provided in Appendix \ref{app:a_s}.

The synthesized program ${\cal{P}}$ serves dual objectives: 1) as an algorithmic solver for task $\tau$, and 2) as a structured data generator for creating reasoning thoughts. Through iterative execution on instances set $\tilde{\mathcal{D}_\tau}$, where $|\tilde{\mathcal{D}_\tau}|=n$, ${\cal{P}}$ produces structured reasoning thoughts set ${T} = \{{T}_i\}_{i=1}^n$ that form the training corpus for LLM fine-tuning.

\begin{algorithm}[hbt]
    \caption{Thought-Enhanced LLM Fine-Tuning}
    \begin{algorithmic}[1]\label{algo:g_t_d}
        \REQUIRE Target task $\tau$, problem instances $\tilde{\mathcal{D}_\tau}$,  $\mathbb{A}$, $\mathbb{S}$, foundation model $\tilde{\text{LLM}}$;
        \STATE $(A, S) \gets {\textbf{Selector}}(\tau, \mathbb{A}, \mathbb{S})$;\COMMENT{Action/State set selection}
        \STATE ${\cal{P}} \gets {\textbf{Constructor}}(\tau, A, S)$; \COMMENT{Program synthesis}
        \STATE Initialize thought corpus ${T} \leftarrow \emptyset$;
        \FOR{$i = 1$ \text{ to } $|\tilde{\mathcal{D}_\tau}|$}
            \STATE Select an instance ${\cal{I}}_i \gets \tilde{\mathcal{D}_\tau}[i]$; 
            \STATE Generate thought ${T}_i \leftarrow {\cal{P}}({\cal{I}}_i)$;\COMMENT{Program execution}
            \STATE ${T} \leftarrow {T} \cup\{{T}_i\}$;
        \ENDFOR
        \STATE \textbf{return} ${\hbox{{LLM}}}_{\text{fine-tuned}} \leftarrow \mathcal{F}(\tilde{\text{LLM}}, {T})$.
    \end{algorithmic}
\end{algorithm}


To achieve the functionalities of {\bf{Selector}} and {\bf{Constructor}}, we propose {\bf Meta-Thought Programming (MTP)}, a systematic methodology for generating $(A, S, {\cal{P}})$ triples through structured knowledge extraction. For GCO problems, we develop two MTP frameworks as follows:

\noindent - {\bf{Forward MTP}}: decomposes classical heuristic algorithms to extract fundamental reasoning patterns through constructive forward-chaining;

\noindent - {\bf{Backward MTP}}: discovers implicit reasoning principles via backward analysis of high-quality solutions.



The GraphThought framework, designed to address the OTD problem, embodies the dual-process framework depicted  in Figure \ref{fig:framework}. The proposed framework consists of two independent components: the forward framework illustrated in the upper half of the figure, whereas the backward one is correspondingly demonstrated in the lower half. The architectural components will be discussed in the following two subsections, collectively establishing the OTD solution set.

\subsection{Forward MTP Framework}\label{sec:forward}

The \textcolor{black}{green-highlighted modules} in Figure \ref{fig:framework} formally establish the forward MTP framework. For a given GCO task $\tau$, this framework systematically extracts several classical heuristic algorithms. These algorithms are systematically combined to distill fundamental operations from $\mathbb{A}$, thereby constructing the action thought set $A$. Concurrently, the corresponding state thought set $S$ is axiomatically derived from $\mathbb{S}$ with $A$ and $\tau$. Following this procedural logic, the target program $\mathcal{P}$ is synthesized via categorical composition of selected $A$-$S$ pairs. The complete generation mechanisms of $A$ and $S$ is rigorously detailed in Appendix \ref{app:a_s_method}.

A basic program template of $\mathcal{P}$ is structured in Program Template \ref{algo:template_heuristic}. It sequentially applies action thoughts in $A$ followed by displaying each state of $S$. This template may require domain-specific adaptations depending on the task $\tau$.
An application of the forward MTP framework for the connected components problem is introduced in Appendix \ref{sec:forward_c_c}. 

\begin{template}[htbp]
\caption{ A Forward MTP Program Template}\label{algo:template_heuristic}
\begin{algorithmic}[1]
\REQUIRE Instance $I$,  state and action thought set $S,A$.
\STATE Initialize an empty solution $x \leftarrow \emptyset$
\STATE Initialize a solving flag $flag \leftarrow False$
\WHILE{$flag$}
    \FOR{$a \in A$}
        \STATE Update solution $x$ according to $a$; \COMMENT{Action thought application with heuristic methods}
        \FOR{$s \in S$}
            \STATE Display state $s$; \COMMENT{State thought application}
        \ENDFOR
        \STATE Update instance $I$;
    \ENDFOR
    \STATE Update $flag$; \COMMENT{ Update the iteration condition}
\ENDWHILE
\end{algorithmic}
\end{template}

\subsection{Backward MTP Framework}

High-quality GCO solutions often embed intelligent problem solving strategies. Although some established knowledge has guided heuristic and approximation algorithm design, many complex patterns remain undiscovered due to analytical complexity. The representational capacity of deep networks enables them to encode such optimal solution patterns, making it promising to generate thoughts for LLMs using approximation or optimal solvers' guidance.

The \textcolor{black}{red-highlighted modules} in Figure \ref{fig:framework} structurally define the backward MTP framework. This architecture employs high-quality solvers (including both exact and approximate methods) to obtain high-quality solutions from which we can extract solution construction patterns. The action thought set is constrained to a single operator, AddOne, which incrementally incorporates solution elements into the current partial solution, thereby generating stepwise trajectories. The state design methodology maintains consistency with the forward framework in Section \ref{sec:forward}. The program $\mathcal{P}$ integrates a task-specific solver $\mathcal{X}$ with the thought-generation process through the Program Template \ref{algo:template_opt}.
An application of the backward MTP framework for the MIS problem is shown in Appendix \ref{sec:backward_mis}. 
\begin{template}[htbp]
\caption{ A Backward MTP Program Template}\label{algo:template_opt}
\begin{algorithmic}[1]
\REQUIRE Instance $I$, action thought set $A$, state thought set $S$, a solver $\mathcal{X}$ of $\tau$.
\STATE Initialize an optimal or suboptimal solution $\hat{x} \leftarrow \mathcal{X}(I)$
\STATE Initialize an empty solution $x \leftarrow \emptyset$
\STATE Initialize a solving flag $flag \leftarrow False$
\WHILE{$flag$}
    \STATE $e \leftarrow \text{AddOne}(\hat{x})$;\COMMENT{Add element from standard solution}
    \STATE $x \leftarrow x \cup \{e\}$;
    \FOR{$s \in S$}
        \STATE Display state $s$; \COMMENT{State thought application}
    \ENDFOR
    \STATE Update instance $I$;
    \STATE Update $flag$;\COMMENT{Update the iteration condition}
\ENDWHILE
\end{algorithmic}
\end{template}

\section{Experiments}\label{sec:expe}

With generated reasoning datasets, we fine-tuned \texttt{Meta-Llama-3-8B-Instruct}\footnote{\url{https://huggingface.co/meta-llama/Meta-Llama-3-8B-Instruct}} using Low-Rank Adaptation (LoRA) via the \texttt{llama-factory} framework\footnote{\url{https://github.com/hiyouga/LLaMA-Factory}}, naming the resulting model {\bf{Llama-GT}} (GT means GraphThought). For inference, we use \texttt{vLLM} \footnote{\url{https://github.com/vllm-project/vllm}} to accelerate. Full hyperparameters are placed in Appendix \ref{app:setting}.
For performance evaluation, we adopted the \textit{optimality} metric from the \texttt{GraphArena} benchmark~\cite{tang2025grapharena}. 
Regarding the definitions of small/large graphs, we strictly followed the specifications in \texttt{GraphArena}: 
i) {\bf{Neighbor/Distance Task}}: Small (contains $4$-$19$ nodes), Large (contains $20$-$50$ nodes);
ii) {\bf{Connected/Diameter/MCP/MIS/MVC Task}}: Small (contains $4$-$14$ nodes), Large (contains $15$-$30$ nodes);
iii) {\bf{MCS/GED/TSP Task}}: Small (contains $4$-$9$ nodes), Large (contains $10$-$20$ nodes). 

The evaluation covered all ten predefined tasks in \texttt{GraphArena}, with detailed task descriptions and optimality criteria provided in Appendix \ref{app:evaluation_dataset}.


\begin{table*}[htb]
\label{tab:example}
\centering
\caption{Performance comparison of optimal solution rates (\%) across $10$ graph tasks, evaluated on small and large graphs, each has $500$ instances. Results are shown for: (1) Code-augmented models (DeepSeek-V2-Coder, GPT-4o-Coder) and Supervised Fine-Tuned Model provided by GraphArena (Qwen2-7B-SFT);(2) Few-shot thought prompting variants (Deepseek-V3, Llama-3.3-70B, etc.); (3) STaR framework implementations with different fewshot strategies; and (4) Llama-GT trained with/without thought mechanisms. Metrics include polynomial tasks (Neighbor, Distance, Connected, Diameter), NP-hard tasks (MVC, MIS, MCP, TSP, MCS, GED). $^*$ indicates that the model in the GraphArena was trained on fewer data compared to the data used in our model.}
\renewcommand{\arraystretch}{1.2} 

\resizebox{0.95\textwidth}{!}{%
\begin{tabular}{lCCCCCCCCCC}

\hline
\multicolumn{11}{c}{Graph Task (Small Graphs)} \\ \hline
Model & Neighbor & Distance & Connected & Diameter & MVC & MIS & MCP & TSP & MCS & GED \\ \hline
DeepSeek-V2-Coder & 0.816 & 0.894 & 0.586 & 0.142 & 0.176 & 0.482 & 0.498 & 0.276 & 0.228 & 0.214 \\ 
GPT-4o-Coder & 0.808 & 0.654 & 0.712 & 0.334 & 0.296 & 0.530 & 0.644 & \textbf{0.490} & 0.508 & 0.320 \\ 
Qwen2-7b-SFT$^*$ & 0.966 & 0.912 & 0.888 & 0.608 & 0.548 & 0.702 & 0.696 & 0.368 & 0.000 & 0.054 \\ 
\hline
+Few-shot Thought & & & & & & & & & & \\ 
Deepseek-V3 & \textbf{1.000} & 0.988 & \textbf{1.000} & 0.850 & 0.366 & 0.642 & 0.754 & 0.370 & \textbf{0.544} & 0.308 \\ 
Llama-3.3-70B & 0.970 & 0.970 & 0.954 & 0.674 & 0.206 & 0.648 & 0.554 & 0.292 & 0.484 & 0.326 \\ 
Mixtral-8x7b & 0.580 & 0.536 & 0.378 & 0.156 & 0.148 & 0.326 & 0.490 & 0.168 & 0.196 & 0.262 \\ 
Llama3-8b-Instruct & 0.700 & 0.480 & 0.502 & 0.070 & 0.108 & 0.248 & 0.258 & 0.190 & 0.222 & 0.450 \\ 
\hline
+SFT & & & & & & & & & & \\ 
STaR(w/ GT) & 0.648 & 0.910 & 0.522 & 0.466 & 0.722 & 0.640 & 0.676 & 0.352 & 0.308 & 0.370 \\ 
STaR(w/ LLM) & 0.296 & 0.662 & 0.440 & 0.380 & 0.688 & 0.654 & 0.288 & 0.282 & 0.328 & 0.258 \\ 
\textbf{Llama-GT(w/o Thought)} & 0.988 & 0.990 & 0.906 & 0.820 & 0.930 & 0.972 & 0.906 & 0.366 & 0.538 & \textbf{0.608} \\ 
\textbf{Llama-GT} & \textbf{1.000} & \textbf{1.000} & 0.996 & \textbf{0.954} & \textbf{0.972} & \textbf{0.994} & \textbf{0.952} & 0.392 & 0.496 & 0.460 \\ 
\hline
\multicolumn{11}{c}{Graph Task (Large Graphs)} \\ \hline
Model & Neighbor & Distance & Connected & Diameter & MVC & MIS & MCP & TSP & MCS & GED \\ \hline
DeepSeek-V2-Coder & 0.672 & 0.632 & 0.206 & 0.008 & 0.080 & 0.028 & 0.072 & 0.000 & 0.022 & 0.014 \\ 
GPT-4o-Coder & 0.868 & 0.684 & 0.378 & 0.112 & 0.110 & 0.072 & 0.222 & 0.028 & \textbf{0.036} & 0.018 \\ 
Qwen2-7b-SFT$^*$ & 0.790 & 0.570 & 0.230 & 0.092 & 0.156 & 0.054 & 0.136 & 0.000 & 0.000 & 0.032 \\ 
\hline
+Few-shot Thought & & & & & & & & & & \\ 
DeepSeek-V3 & \textbf{0.992} & 0.942 & \textbf{0.932} & 0.448 & 0.120 & 0.304 & 0.290 & 0.020 & 0.012 & 0.020 \\ 
Llama-3.3-70B & 0.952 & 0.866 & 0.856 & 0.290 & 0.118 & 0.254 & 0.136 & 0.000 & 0.012 & 0.012 \\ 
Mixtral-8x7b & 0.466 & 0.282 & 0.108 & 0.002 & 0.032 & 0.024 & 0.056 & 0.000 & 0.000 & 0.022 \\ 
Llama3-8b-Instruct & 0.604 & 0.220 & 0.132 & 0.002 & 0.028 & 0.010 & 0.038 & 0.000 & 0.002 & 0.026 \\ 
\hline
+SFT & & & & & & & & & & \\ 
STaR(w/ GT) & 0.374 & 0.618 & 0.124 & 0.080 & 0.360 & 0.216 & 0.134 & 0.018 & 0.006 & 0.026 \\ 
STaR(w/ LLM) & 0.320 & 0.332 & 0.090 & 0.038 & 0.126 & 0.124 & 0.030 & 0.002 & 0.004 & 0.024 \\ 
\textbf{Llama-GT(w/o Thought)} & 0.804 & 0.864 & 0.340 & 0.302 & 0.652 & 0.652 & 0.370 & 0.024 & 0.020 & \textbf{0.068} \\ 
\textbf{Llama-GT} & 0.988 & \textbf{0.984} & 0.836 & \textbf{0.600} & \textbf{0.744} & \textbf{0.900} & \textbf{0.634} & \textbf{0.036} & \textbf{0.036} & 0.008 \\ 
\hline

\end{tabular}

}
\end{table*}

\noindent \textbf{Comparison with Original GraphArena}.
We present the majority of GraphArena's original results in Appendix~\ref{app:complete_main_result}, retaining only the key comparisons with DeepSeek-V2-Coder, GPT-4o-Coder, and Qwen2-7B-SFT. 
While code-augmented approaches (DeepSeek-V2-Coder and GPT-4o-Coder) demonstrate intermediate performance on fundamental tasks, they exhibit limitations when confronted with complex challenges such as graph diameter computation ($0.334$ optimality rate), which Llama-GT effectively addresses ($0.954$). 
Notably, Qwen2-7B-SFT achieves strong performance on basic operations but fails catastrophically on MIS for large graphs ($0.054$), underscoring the constraints of supervised fine-tuning without a thought-enhanced paradigm.

\noindent \textbf{Comparison with Few-shot Thought Prompting}.
While few-shot prompting improves performance, Llama-GT shows greater consistency, especially on large graphs. 
DeepSeek-V3 achieves perfect scores on simple tasks like Neighbor ($1.0$) but struggles with MVC ($0.366$). 
Llama-GT outperforms DeepSeek-V3 on NP-hard tasks like MVC ($0.744$ vs $0.120$) and MIS ($0.900$ vs $0.304$), demonstrating the superiority of the GraphThought framework over few-shot exemplars.

\noindent \textbf{Comparison with STaR Framework}.
Comparison with the STaR~\cite{zelikman2022star} framework's self-generated thought approach reveals critical insights. 
Using STaR with Llama-3-8B-Instruct under two configurations\footnote{More detailed configuations see appendix \ref{app:setting}.}: 
(1) few-shot prompting with thoughts generated with our GraphThought framework and 
(2) with \texttt{LLM}-generated thoughts.
Llama-GT substantially outperforms both variants across all graph sizes, showing $2$-$3$x improvements on large graphs (MVC: $0.744$ vs $0.360$, MIS: $0.900$ vs $0.216$). 
The gap persists even when STaR uses our thought-enhanced examples, emphasizing the limits of iterative self-improvement approaches. 
In addition, STaR(w/GT) performs better than STaR(w/LLM) across nearly all tasks and different graph sizes, further underscoring the importance of our thought paradigm over automated generation for training.

\noindent \textbf{Ablation of Thought Mechanism Impact}
The thought mechanism significantly improves performance on NP-hard tasks. 
Llama-GT without thought integration still performs well on polynomial tasks (e.g., Neighbor: $0.988$ small, $0.804$ large). 
With thought integration, MVC increases from $0.930$ to $0.972$ (small) and $0.652$ to $0.744$ (large), MIS from $0.972$ to $0.994$ (small) and $0.652$ to $0.900$ (large). 
However, performance drops on GED (small: $0.608$ to $0.460$; large: $0.068$ to $0.008$). 
We posit that potential causes underlying this phenomenon will be explored in the later discussion.

\begin{figure}[htb]
\begin{center}
\includegraphics[width=\columnwidth]{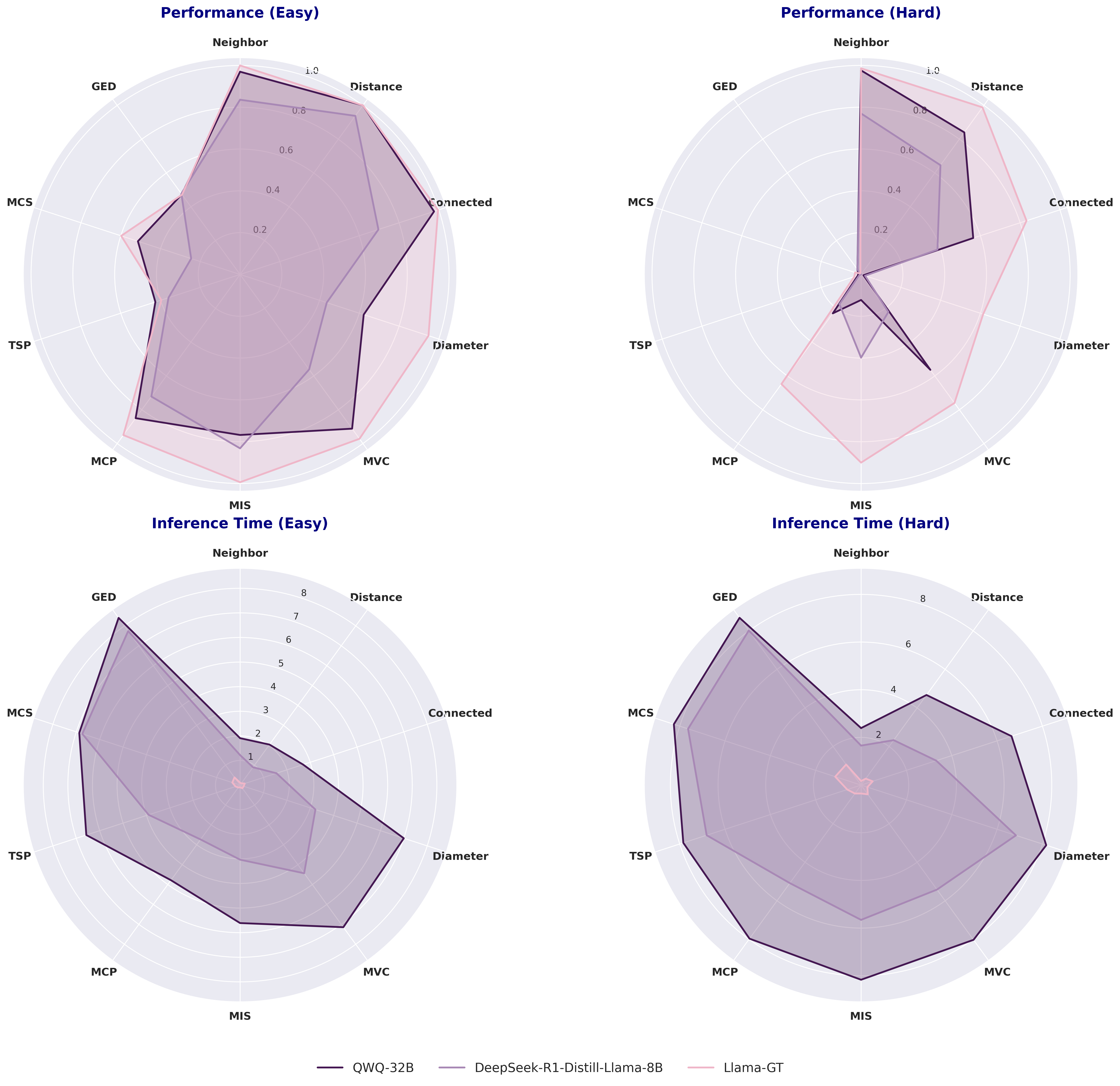} \\
\caption{Performance and time cost of inference model and Llama-GT on ten graph tasks of GraphArena benchmark. (Upper-Left) Performance on small graph instances. (Upper-Right) Performance on large ones. (Bottom-Left) Inference time cost on small ones. (Bottom-Right) Inference time cost on large ones.}
\label{fig:vs_inference_model}
\end{center}
\vspace{-0.5cm}
\end{figure}

\noindent \textbf{Performance Comparison on Reasoning}.
We conducted a comprehensive performance comparison between Llama-GT and two state-of-the-art inference models: QWQ-32B and DeepSeek-R1-Distill-Llama-8B. All models were evaluated on 500 instances per task using vLLM with identical experimental configurations. As demonstrated in Figure~\ref{fig:vs_inference_model}, Llama-GT achieves comparable or superior performance to both baseline models across polynomial-time tasks and specific NP-hard problems, particularly excelling in MVC and MIS tasks.

Our analysis reveals a critical efficiency advantage: conventional inference models exhibit prolonged cognitive processing phases during problem-solving, often generating redundant computational steps that exceed the maximum token threshold, resulting in significantly slower inference speeds. In contrast, our trained model constrains these cognitive mechanisms to operate within fixed patterns, achieving substantial time efficiency improvements while maintaining solution quality as shown in Figure \ref{fig:vs_inference_model}.

However, performance disparities persist in more complex CO tasks such as TSP, Graph Edit Distance (GED), and Maximum Common Subgraph (MCS). We provide complete quantitative results and a systematic analysis of these performance degradation patterns in Appendix~\ref{app:performace_degradation}.


\noindent \textbf{BoN-Enhanced Optimality Rates versus Heuristics}.
We introduce a generalized optimality ratio metric for evaluating the quality of a found solution given an instance. 
For each problem instance \( I \), the optimal solution \( x^* \) is obtained using Gurobi, which guarantees exact optimality. 
The optimality ratio for a solution \( x \) on the instance \( I \) of a GCO problem \( \tau \) is formally defined as:
\begin{equation}
{\textstyle{\text{Rate}(I,x) = \min \left\{ \frac{\phi(x^*)}{\phi(x)}, \frac{\phi(x)}{\phi(x^*)} \right\},}}
\end{equation}
where \( \phi(\cdot) \) is the evaluation function of the solution of \( \tau \). 
This symmetric ratio works for both maximization and minimization tasks. 
For maximization tasks (e.g., MIS), where \( \phi(x) \leq \phi(x^*) \), the ratio \( \frac{\phi(x)}{\phi(x^*)} \) directly measures approximation quality within \([0,1]\). 
For minimization problems (e.g., MVC), the inverse ratio \( \frac{\phi(x^*)}{\phi(x)} \) appropriately penalizes suboptimal solutions while maintaining the same normalized range. 
This metric enables cross-problem performance comparisons while maintaining the interpretability of near-optimal solutions.

\begin{figure}[htbp]
\begin{center}
\centerline{\includegraphics[width=\columnwidth]{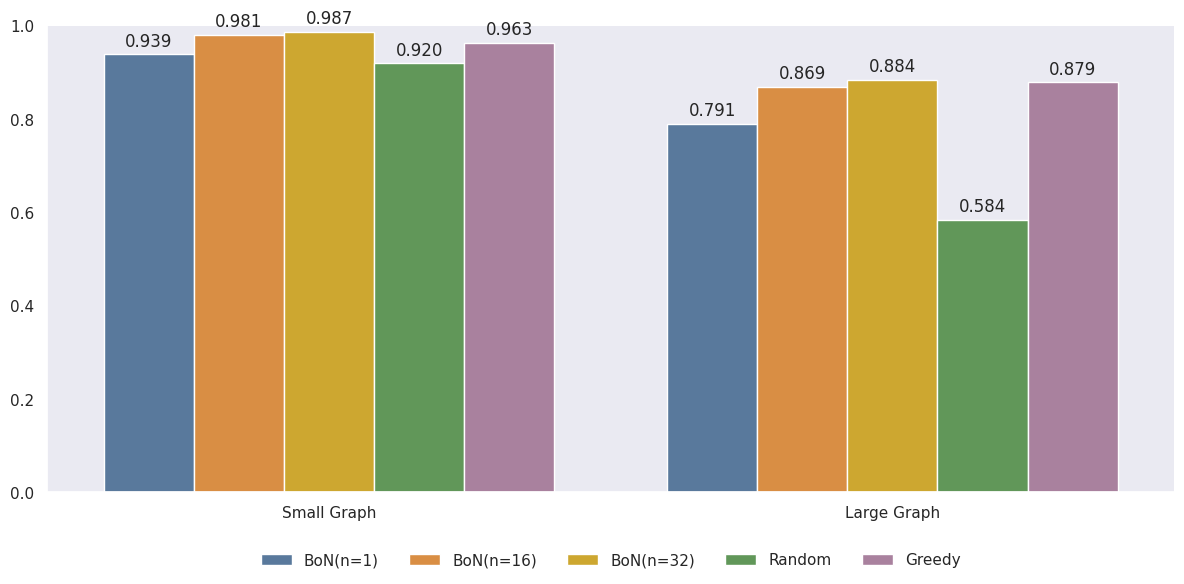}}
\caption{Average performance of Llama-GT with BoN strategy and classic solvers on $6$ NP-Hard Tasks (GED, MCP, MCS, MIS, MVC, TSP) for small and large graphs. The values represent the average optimality ratio across different tasks.}
\label{fig:vs_heuristic}
\end{center}
\vspace{-0.5cm}
\end{figure}

Figure \ref{fig:vs_heuristic} demonstrates the average optimality ratio through comprehensive evaluations across $1,000$ graph instances ($500$ small and $500$ large) spanning $6$ NP-Hard tasks: GED, MCP, MCS, MIS, MVC and TSP. 
Our base model (\( n=1 \), without Best-of-\( N \) enhancement) achieves remarkable optimality ratios of $93.9$\% on small graphs and $79.1$\% on large graphs, outperforming random baselines by significant margins of $1.9$\% and $20.7$\% respectively. 
These results demonstrate the inherent effectiveness of our Llama-GT model in generalizing across problem scales and types.


We enhance solution quality with the Best-of-N (BoN) strategy, which generates \( n \) candidate solutions per instance and selects the best. 
With BoN(n=32), optimality improves to 98.7\% on small graphs and $88.4$\% on large graphs, surpassing greedy algorithms ($96.3$\% and $87.9$\%, respectively). 
This technique significantly narrows the performance gap between data-driven approaches and manual heuristics, achieving near-optimal results without problem-specific rules.

A case study in the MIS problem (Figure \ref{fig:greedy_case}) highlights our approach’s superiority. 
While the greedy algorithm selects low-degree nodes (e.g., node $16$), our method uses adaptive heuristics to prioritize high-impact nodes (e.g., node $18$) and optimize selections based on evolving graph structures, achieving a $12$-node MIS compared to the greedy algorithm’s $11$. 
This demonstrates how our framework avoids the myopic decisions of traditional heuristics, leading to higher-quality solutions.

\section{Conclusion}

We introduce \textbf{GraphThought}, a structured framework for improving the reasoning capabilities of LLM on GCO problems. By formalizing the OTD problem, we enable the construction of high-quality reasoning trajectories via two distinct data generation strategies: a heuristic-driven forward method for polynomial-time tasks, and a solver-aligned backward method for NP-hard problems. Fine-tuning LLMs with these structured trajectories yields Llama-GT, a compact 8B-parameter model that achieves state-of-the-art performance on the GraphArena benchmark, surpassing significantly larger models across multiple GCO tasks. These results suggest that incorporating structured reasoning into LLM training holds substantial potential for advancing CO with language-based models.

\section{Limitations}
\label{sec:limitations}
Despite the promising results achieved by GraphThought framework, several important limitations remain, which are outlined below.

\paragraph{Performance on Complex or Knowledge-Scarce Problems.}
While the proposed framework significantly enhances the reasoning capabilities of large language models (LLMs) through structured reasoning, its performance deteriorates when applied to problems of significantly higher complexity or those lacking explicit combinatorial priors—such as the Traveling Salesman Problem (TSP), Graph Edit Distance (GED), and Maximum Common Subgraph (MCS)—as discussed in Section~\ref{app:performace_degradation}. These challenges are particularly pronounced in NP-hard graph optimization tasks that require long reasoning chains or lack clear heuristic guidance. 

\paragraph{Efficiency Concerns of LLM-Based Solvers.}
Employing large language models as combinatorial problem solvers inherently introduces computational inefficiencies. Compared to traditional optimization solvers like Gurobi or specialized heuristic algorithms, LLM-based approaches suffer from higher inference latency and substantial computational resource demands. This inefficiency limits their practical deployment, particularly in scenarios requiring rapid or real-time responses.

\paragraph{Difficulty in Scaling to Large Graph Instances.}
A crucial limitation arises from the inability of our method to efficiently address large-scale graph instances. Due to the verbose and sequential nature of natural language reasoning employed by LLMs, handling large graphs—comprising thousands or millions of nodes—poses significant representation and computational challenges. These large graph instances typically exceed the context length of current LLM architectures, resulting in fragmented or incomplete reasoning processes and thus severely compromising solution quality and validity.

In summary, while our GraphThought framework demonstrates substantial improvements in LLM capabilities for GCO, addressing the aforementioned limitations remains a vital direction for future research.

\clearpage
\newpage


\bibliography{main}

\newpage


\include{appendix}

\end{document}

%% file: appendix.tex

\appendix
\onecolumn

\section{Related Work}

\subsection{LLMs for Graph Combinatorial Optimization}

The integration of Large Language Models (LLMs) into solving Graph Combinatorial Optimization (GCO) problems has garnered significant attention in recent years. Several works have attempted to leverage LLMs for GCO problems from various perspectives. Jin et al.~\citep{jin2024large} provided a comprehensive survey on the application of LLMs on graphs, categorizing potential scenarios into pure graphs, text-attributed graphs, and text-paired graphs. They discussed techniques such as using LLMs as predictors, encoders, and aligners. Chen et al.~\citep{chen2024exploring} explored the potential of LLMs in graph machine learning, especially for node classification tasks, and investigated two pipelines: LLMs-as-Enhancers and LLMs-as-Predictors. Liu et al.~\citep{liu2023towards} introduced the concept of Graph Foundation Models (GFMs) and classified existing work into categories based on their dependence on graph neural networks and LLMs. Ren et al.~\citep{ren2024survey} conducted a survey on LLMs for graphs, proposing a taxonomy to categorize existing methods based on their framework design.

\subsubsection{Prompt Engineering and Benchmark Development}
Recent advances in LLM applications for graph problems have focused on prompt engineering strategies and comprehensive benchmarking frameworks. Wang et al.~\citep{wang2024can} introduced the NLGraph benchmark with Build-a-Graph prompting and algorithmic optimizations for graph reasoning tasks. Tang et al.~\citep{tang2025grapharena} developed GraphArena as a standardized benchmarking platform, while Yuan et al.~\citep{yuan2024gracore} proposed GraCoRe for assessing graph comprehension capabilities. Skianis et al.~\citep{skianis2024graph} demonstrated the effectiveness of pseudo-code prompting for enhancing graph problem performance.

\subsubsection{Architectural and Framework Innovations}
Several architectural innovations have emerged to enhance LLM capabilities in graph processing. Li et al.~\citep{li2024graphteam} introduced GraphTeam as a multi-agent collaborative system, complemented by Hu et al.'s GraphAgent-Reasoner framework~\citep{hu2024scalable}. Perozzi et al.~\citep{perozzi2024let} developed GraphToken for explicit structured data representation, while Cao et al.~\citep{cao2024graphinsight} proposed GraphInsight to improve structural comprehension.

\subsubsection{Graph Representation and Encoding Strategies}
Innovative graph encoding methods have been crucial for bridging the gap between LLMs and graph structures. Fatemi et al.~\citep{fatemi2023talk} pioneered text-based graph encoding, while Elhenawy et al.~\citep{elhenawy2024eyeballing, elhenawy2024visual} explored multimodal visual reasoning for TSP solutions. Feng et al.~\citep{feng2024beyond} extended these approaches to hypergraphs through LLM4Hypergraph.

\subsubsection{Empirical Analysis and Performance Factors}
Comprehensive empirical studies have revealed critical insights into LLM capabilities and limitations. Guo et al.~\citep{guo2023gpt4graph} conducted large-scale evaluations on graph-structured data, while Wang et al.~\citep{wang2024reasoning} analyzed temperature's impact on reasoning performance. Firooz et al.~\citep{firooz2024lost} investigated contextual proximity effects, and Dai et al.~\citep{dai2024revisiting} provided case studies on graph reasoning limitations. Sanford et al.~\citep{sanford2024understanding} analyzed transformer architectures through graph algorithmic lenses.

Other works have focused on training graph foundation models using dense embeddings from LLM pretraining. Drakulic et al.~\citep{drakulic2024goal} proposed GOAL, a generalist model for solving multiple combinatorial optimization problems. Jiang et al.~\citep{jiang2024unco} introduced UNCO, a unified framework for solving various COPs using LLMs. Anonymous~\citep{anonymous2024solving} proposed a unified model for diverse CO problems using a transformer backbone. Chai et al.~\citep{chai2023graphllm} introduced GraphLLM to boost the graph reasoning ability of LLMs. Wei et al.~\citep{wei2024rendering} proposed GITA, a framework integrating visual and textual information for graph reasoning.

\subsection{Chain-of-Thought, Tree-of-Thought, and Graph-of-Thought Methods}
The Chain-of-Thought (CoT) prompting technique, introduced by Wei et al. \cite{wei2022chain}, demonstrates that generating intermediate reasoning steps can significantly improve LLMs' performance on complex reasoning tasks. This method was further extended by Yao et al. \cite{yao2024tree}, who proposed the Tree of Thoughts (ToT) framework, enabling exploration over coherent units of text (thoughts) to enhance problem-solving abilities. The ToT approach allows LMs to consider multiple reasoning paths and self-evaluate choices, significantly improving performance on tasks requiring non-trivial planning or search. Besta et al. \cite{besta2024graph} introduced the Graph of Thoughts (GoT), which advances prompting capabilities by modeling LLM-generated information as an arbitrary graph. This approach enables combining arbitrary LLM thoughts into synergistic outcomes and enhances thoughts using feedback loops. Besta et al. \cite{besta2024demystifying} further demystified the concepts of chains, trees, and graphs of thoughts, providing a taxonomy of structure-enhanced LLM reasoning schemes. These studies highlight the importance of structured reasoning topologies in improving LLMs' problem-solving abilities.

\subsection{Self-Correction, Planning, and Search Strategy Learning}
Huang et al. \cite{huang2024large} examined the role of self-correction in LLMs, finding that intrinsic self-correction without external feedback often fails to improve reasoning accuracy. In contrast, Lehnert et al. \cite{searchformer} proposed the Searchformer model, which predicts the search dynamics of the A* algorithm, significantly outperforming traditional planners on complex decision-making tasks. Gandhi et al. \cite{gandhi2024stream} introduced the Stream of Search (SoS) approach, teaching language models to search by representing the process as a flattened string. This method significantly improved search accuracy and enabled flexible use of different search strategies.
    
\section{Experiments Setting}
\label{app:setting}

\subsection{Training Parameter}

Our framework is implemented based on LLaMA-Factory. We perform supervised fine-tuning (SFT) on the Meta-Llama-3-8B-Instruct\footnote{\url{https://huggingface.co/meta-llama/Meta-Llama-3-8B-Instruct}} model using LoRA adaptation. The training process is conducted on a NVIDIA H800 PCIe 80GB GPU with 30K instruction-following examples. Key hyperparameters include a cosine learning rate scheduler with initial value 8e-4, 4 training epochs, and gradient accumulation over 8 steps. The complete configuration details are presented in Table~\ref{tab:training_config}.

\begin{table}[h]
\centering
\caption{Training Configuration}
\label{tab:training_config}
\begin{tabular}{ll}
\toprule
\textbf{Category} & \textbf{Setting} \\
\midrule
\textbf{Model Configuration} & \\
\quad Base Model & Llama-3-8B-Instruct \\
\quad Fine-tuning Method & LoRA \\
\quad Hardware & a NVIDIA H800 PCIe 80GB GPU \\
\quad Dataset Size & 30,000 samples \\
\quad Validation Split & 10\% \\
\quad Max Sequence Length & 3,000 tokens \\
\midrule
\textbf{Training Parameters} & \\
\quad Epochs & 4 \\
\quad Learning Rate & 8e-4 \\
\quad Batch Size (per device) & 8 \\
\quad Gradient Accumulation Steps & 8 \\
\quad Optimizer & AdamW \\
\quad Learning Rate Scheduler & Cosine \\
\quad Warmup Steps & 0 \\
\quad Max Gradient Norm & 1.0 \\
\midrule
\textbf{LoRA Configuration} & \\
\quad LoRA Rank & 8 \\
\quad LoRA Alpha & 16 \\ 
\quad LoRA Dropout & 0 \\
\bottomrule
\end{tabular}
\end{table}

We follow the dataset construction methodology described in the original STaR\cite{zelikman2022star} paper. Specifically, we first perform inference using either an untrained or partially trained model, then conduct rationalization on unsuccessful cases that failed to reach the optimal solution. After collecting these rationalized examples to construct the training dataset, we subsequently perform supervised fine-tuning (SFT) on the base model. For the base model, we consistently employ Llama-3-8B-Instruct with identical hyperparameter settings as listed in the configuration Table~\ref{tab:training_config}. The complete STaR process undergoes four iterative cycles to progressively enhance model performance.

\subsection{Inference Setting}

During the inference phase, we deployed the base untrained and trained model using the vllm framework, including Llama-GT, the original Meta-Llama-3-8B-Instruct model and the reasoning models (QWQ-32B\footnote{\url{https://huggingface.co/Qwen/QwQ-32B}} and DeepSeek-R1-Distill-Llama-8B\footnote{\url{https://huggingface.co/deepseek-ai/DeepSeek-R1-Distill-Llama-8B}}).
Other models utilized API-based inference (including Deepseek-V3,\footnote{\url{https://api.deepseek.com}}, llama-3.3-70b, mixtral-8x7b and QwQ-32B-Preview\footnote{\url{https://api.pandalla.ai/}}). 
To ensure reproducibility, we configured the temperature parameter to 0.1 for all models. Due to computational constraints, each API-based model underwent single-pass inference for each test instance.

For Best-of-N(BoN) experiments investigating whether increased inference-time computation enhances model performance, we adjusted the batch size to 16 or 32 while maintaining a temperature setting of 1.0 to maximize response diversity.

\subsection{Evaluation Dataset}
\label{app:evaluation_dataset}

We conduct evaluations using GraphArena's original test datasets with preserved graph size definitions ("small" vs "large") from their benchmark. The 10 tasks are categorized by time complexity and specified as follows:

\begin{itemize}[leftmargin=*]
    \item \textbf{Common Neighbor (Polynomial)}: For graph $\mathcal{G} = \{\mathcal{V}, \mathcal{E}\}$ and nodes $v_1, v_2$, identify all $u \in \mathcal{V}$ connecting to both. Optimal solution maximizes $|S|$ where $S = \{u \mid (u,v_1),(u,v_2) \in \mathcal{E}\}$. {\bf Neighbor} for short.
    
    \item \textbf{Shortest Distance (Polynomial)}: Find the path between $v_1$ and $v_2$ with minimal hops in $\mathcal{G}$. Optimality requires $\min \ell(p_{v_1 \to v_2})$. {\bf Distance} for short.
    
    \item \textbf{Connected Component (Polynomial)}: Select representative nodes covering all components. Optimality demands full coverage ($|\mathcal{C}| =$ total components). {\bf Connected} for short.
    
    \item \textbf{Graph Diameter (Polynomial)}: Identify the longest shortest path. The optimal solution achieves $\max_{u,v} d(u,v)$. {\bf Diameter} for short.
    
    \item \textbf{Maximum Clique Problem (NP-hard)}: A clique is a complete subgraph where every pair of distinct vertices is connected. The task requires identifying the largest such subgraph $\mathcal{C} \subseteq \mathcal{V}$ in $\mathcal{G}$. A solution is optimal if no larger clique exists in the graph (i.e., $\nexists \mathcal{C}'$ where $|\mathcal{C}'| > |\mathcal{C}|$). {\bf MCP} for short.
    
    \item \textbf{Maximum Independent Set (NP-hard)}: Select an independent set $\mathcal{S}$. Optimality requires identifying the maximum-size set of mutually non-adjacent nodes. {\bf MIS} for short.
    
    \item \textbf{Minimum Vertex Cover (NP-hard)}: Determine a vertex cover $\mathcal{S}$. Optimal solution satisfies covering all edges with the minimum number of nodes. {\bf MVC} for short.
    
    \item \textbf{Maximum Common Subgraph (NP-hard)}: Identify the largest node-induced subgraph $\mathcal{S}$ common to both $\mathcal{G}$ and $\mathcal{H}$ where edge relationships are preserved, with optimality determined by maximizing $|V(\mathcal{S})|$. {\bf MCS} for short.
    
    \item \textbf{Graph Edit Distance (NP-hard)}: Calculate minimal edit operations (node/edge changes) to align $\mathcal{G}$ with $\mathcal{H}$. Optimal solution minimizes total edit cost. {\bf GED} for short.
    
    \item \textbf{Traveling Salesman Problem (NP-hard)}: In a complete graph, find shortest Hamiltonian cycle. Optimal route minimizes $\sum w(e_i)$. {\bf TSP} for short.
\end{itemize}

In alignment with the original study's framework, we maintain the original node range definitions across task categories: (1) Neighbor/Distance tasks operate with small graphs (4-19 nodes) and large graphs (20-50 nodes); (2) Component/Diameter measurements alongside combinatorial problems (MCP/MIS/MVC) utilize small-scale graphs (4-14 nodes) contrasting with large-scale counterparts (15-30 nodes); (3) MCS/GED/TSP adhere to the established size parameters of small (4-9 nodes) versus large (10-20 nodes) instances. By employing the benchmark's test datasets with unmodified size criteria, we preserve experimental continuity and facilitate meaningful cross-study comparisons.

To address more general graph-theoretic challenges and facilitate algorithmic processing, we systematically convert the input representations of these tasks into a unified and structured schema. Detailed descriptions of the input structures and task specifications for each problem are documented in Appendix \ref{app:thought_case}.

\section{License for \textbf{GraphArena} benchmark}

We utilize the \textbf{GraphArena} benchmark to evaluate large language models on graph computational problems solely for academic research purposes. 
GraphArena is open-sourced under the \textbf{Creative Commons Attribution 4.0 International (CC BY 4.0)} license, as indicated in the official publication \cite{tang2025grapharena}.

The datasets incorporated within GraphArena originate from various sources, each with its respective licensing terms:

\begin{itemize}
  \item \textbf{DBLP}: CC0 1.0 Public Domain Dedication
  \item \textbf{Social Network}: CC BY-SA 3.0 License
  \item \textbf{DBpedia}: CC BY-SA 3.0 License
  \item \textbf{OpenFlights}: Database Contents License (DbCL) v1.0
  \item \textbf{PubChemQC (PCQM4Mv2)}: CC BY 4.0 License
\end{itemize}

All datasets are publicly available and have been utilized in accordance with their respective licenses. The datasets employed, including those within the GraphArena benchmark, are either synthetic or derived from publicly available sources with appropriate anonymization measures in place.

\section{Training and Inference Cost}

We report below the training and inference costs for transparency:
\begin{itemize}[leftmargin=*]
    \item \textbf{Training Cost}: The training process for the Llama-3-8B-Instruct model on the GraphThought dataset required approximately 138,828 seconds (38 hours) on a single NVIDIA H800 PCIe 80GB GPU.
    \item \textbf{Inference Cost}: The inference time per instance is approximately 0.0694 seconds for the Llama-3-8B-Instruct model and 0.0703 seconds for the fine-tuned model (Llama-GT). Notably, when using the fine-tuned model with Best-of-N (N=16), the inference time increases to approximately 0.1643 seconds per instance. This increase is not proportional to the number of queries (N=16) due to batch processing, which reduces computational overhead while generating more diverse responses to select the optimal answer.

    \begin{table}[h]
    \centering
    \caption{Inference Time Cost of the Original Model, Llama-GT, and Llama-GT with Best-of-N (N=16)}
    \begin{tabular}{lc}
    \toprule
    \textbf{Model} & \textbf{Time (s)} \\
    \midrule
    Llama-3-8B-Instruct & 0.0694 \\
    Llama-GT & 0.0703 \\
    Llama-GT (N=16) & 0.1643 \\
    \bottomrule
    \end{tabular}
    \label{tab:inference_cost}
    \end{table}
\end{itemize}

\section{Complete Main Result with origin GraphArena results}
\label{app:complete_main_result}
\begin{table*}[!htb]
\label{tab:example}
\centering
\caption{Performance comparison of optimal solution rates (\%) across $10$ graph tasks, evaluated on small and large graphs, each has $500$ instances. Results are shown for: (1) Original inference results of LLMs from GraphArena (Claude3-haiku, GPT-4o, etc.); (2) Code-augmented models (DeepSeek-V2-Coder, GPT-4o-Coder); (3)Supervised Fine-Tuned Model provided by GraphArena (Qwen2-7B-SFT);(4) Few-shot thought prompting variants (Deepseek-V3, Llama-3.3-70B, etc.); (5) STaR framework implementations with different fewshot strategies; and (6) Our Llama-GT model trained with/without thought mechanisms. Metrics include polynomial tasks (Neighbor, Distance, Connected, Diameter), NP-hard tasks (MVC, MIS, MCP, TSP, MCS, GED). $^*$ indicates that the model in the GraphArena paper was trained on fewer data compared to the data used in our model.}
\renewcommand{\arraystretch}{1.2} 

\resizebox{0.95\textwidth}{!}{%
\begin{tabular}{lCCCCCCCCCC}

\hline
\multicolumn{11}{c}{Graph Task (Small Graphs)} \\ \hline
Model & Neighbor & Distance & Connected & Diameter & MVC & MIS & MCP & TSP & MCS & GED \\ \hline
Claude3-haiku & 0.768 & 0.580 & 0.260 & 0.116 & 0.336 & 0.450 & 0.482 & 0.242 & 0.282 & 0.216 \\ 
DeepSeek-V2 & 0.540 & 0.814 & 0.474 & 0.226 & 0.376 & 0.360 & 0.400 & 0.368 & 0.236 & 0.282 \\ 
Gemma-7b & 0.410 & 0.496 & 0.014 & 0.086 & 0.128 & 0.212 & 0.254 & 0.134 & 0.020 & 0.028 \\ 
gpt-3.5-turbo-0125 & 0.562 & 0.572 & 0.096 & 0.130 & 0.376 & 0.184 & 0.400 & 0.230 & 0.176 & 0.144 \\ 
gpt-4o-2024-0513 & 0.860 & 0.796 & 0.794 & 0.426 & 0.326 & 0.518 & 0.528 & 0.404 & 0.406 & 0.268 \\ 
Llama3-70b-Instruct & 0.674 & 0.894 & 0.632 & 0.248 & 0.420 & 0.368 & 0.428 & 0.232 & 0.442 & 0.316 \\ 
Llama3-8b-Instruct & 0.368 & 0.412 & 0.248 & 0.114 & 0.318 & 0.282 & 0.280 & 0.162 & 0.096 & 0.072 \\ 
Mixtral-8x7b & 0.530 & 0.566 & 0.286 & 0.130 & 0.130 & 0.116 & 0.278 & 0.188 & 0.098 & 0.124 \\ 
Qwen1.5-72b-Chat & 0.572 & 0.478 & 0.394 & 0.118 & 0.224 & 0.386 & 0.388 & 0.228 & 0.298 & 0.174 \\ 
Qwen1.5-8b-Chat & 0.138 & 0.266 & 0.022 & 0.048 & 0.138 & 0.118 & 0.216 & 0.170 & 0.062 & 0.058 \\ 
DeepSeek-V2-Coder & 0.816 & 0.894 & 0.586 & 0.142 & 0.176 & 0.482 & 0.498 & 0.276 & 0.228 & 0.214 \\ 
GPT-4o-Coder & 0.808 & 0.654 & 0.712 & 0.334 & 0.296 & 0.530 & 0.644 & \textbf{0.490} & 0.508 & 0.320 \\ 
Qwen2-7b-SFT$^*$ & 0.966 & 0.912 & 0.888 & 0.608 & 0.548 & 0.702 & 0.696 & 0.368 & 0.000 & 0.054 \\ 
\hline
+Few-shot Thought & & & & & & & & & & \\ 
Deepseek-V3 & \textbf{1.000} & 0.988 & \textbf{1.000} & 0.850 & 0.366 & 0.642 & 0.754 & 0.370 & \textbf{0.544} & 0.308 \\ 
Llama-3.3-70B & 0.970 & 0.970 & 0.954 & 0.674 & 0.206 & 0.648 & 0.554 & 0.292 & 0.484 & 0.326 \\ 
Mixtral-8x7b & 0.580 & 0.536 & 0.378 & 0.156 & 0.148 & 0.326 & 0.490 & 0.168 & 0.196 & 0.262 \\ 
QwQ-32B-Preview & 0.962 & 0.848 & 0.696 & 0.514 & 0.262 & 0.482 & 0.560 & 0.310 & 0.444 & 0.318 \\ 
Llama3-8b-Instruct & 0.700 & 0.480 & 0.502 & 0.070 & 0.108 & 0.248 & 0.258 & 0.190 & 0.222 & 0.450 \\ 
\hline
+SFT & & & & & & & & & & \\ 
STaR(w/ GT) & 0.648 & 0.910 & 0.522 & 0.466 & 0.722 & 0.640 & 0.676 & 0.352 & 0.308 & 0.370 \\ 
STaR(w/ LLM) & 0.296 & 0.662 & 0.440 & 0.380 & 0.688 & 0.654 & 0.288 & 0.282 & 0.328 & 0.258 \\ 
\textbf{Llama-GT(w/o Thought)} & 0.988 & 0.990 & 0.906 & 0.820 & 0.930 & 0.972 & 0.906 & 0.366 & 0.538 & \textbf{0.608} \\ 
\textbf{Llama-GT} & \textbf{1.000} & \textbf{1.000} & 0.996 & \textbf{0.954} & \textbf{0.972} & \textbf{0.994} & \textbf{0.952} & 0.392 & 0.496 & 0.460 \\ 
\hline
\multicolumn{11}{c}{Graph Task (Large Graphs)} \\ \hline
Model & Neighbor & Distance & Connected & Diameter & MVC & MIS & MCP & TSP & MCS & GED \\ \hline
Claude3-haiku & 0.406 & 0.358 & 0.052 & 0.002 & 0.076 & 0.014 & 0.090 & 0.000 & 0.000 & 0.018 \\ 
DeepSeek-V2 & 0.278 & 0.534 & 0.154 & 0.022 & 0.064 & 0.032 & 0.032 & 0.006 & 0.000 & 0.014 \\ 
Gemma-7b & 0.116 & 0.246 & 0.002 & 0.004 & 0.014 & 0.006 & 0.016 & 0.000 & 0.000 & 0.002 \\ 
gpt-3.5-turbo-0125 & 0.412 & 0.322 & 0.012 & 0.008 & 0.082 & 0.008 & 0.052 & 0.000 & 0.000 & 0.006 \\ 
gpt-4o-2024-0513 & 0.674 & 0.550 & 0.370 & 0.032 & 0.102 & 0.034 & 0.102 & 0.004 & 0.000 & 0.018 \\ 
Llama3-70b-Instruct & 0.434 & 0.530 & 0.264 & 0.034 & 0.114 & 0.026 & 0.106 & 0.000 & 0.000 & 0.008 \\ 
Llama3-8b-Instruct & 0.118 & 0.234 & 0.022 & 0.002 & 0.064 & 0.010 & 0.022 & 0.000 & 0.000 & 0.002 \\ 
Mixtral-8x7b & 0.232 & 0.282 & 0.040 & 0.000 & 0.036 & 0.008 & 0.034 & 0.000 & 0.000 & 0.006 \\ 
Qwen1.5-72b-Chat & 0.236 & 0.258 & 0.068 & 0.006 & 0.038 & 0.018 & 0.022 & 0.000 & 0.000 & 0.010 \\ 
Qwen1.5-8b-Chat & 0.026 & 0.156 & 0.002 & 0.000 & 0.002 & 0.000 & 0.016 & 0.000 & 0.000 & 0.006 \\ 
DeepSeek-V2-Coder & 0.672 & 0.632 & 0.206 & 0.008 & 0.080 & 0.028 & 0.072 & 0.000 & 0.022 & 0.014 \\ 
GPT-4o-Coder & 0.868 & 0.684 & 0.378 & 0.112 & 0.110 & 0.072 & 0.222 & 0.028 & \textbf{0.036} & 0.018 \\ 
Qwen2-7b-SFT$^*$ & 0.790 & 0.570 & 0.230 & 0.092 & 0.156 & 0.054 & 0.136 & 0.000 & 0.000 & 0.032 \\ 
\hline
+Few-shot Thought & & & & & & & & & & \\ 
DeepSeek-V3 & \textbf{0.992} & 0.942 & \textbf{0.932} & 0.448 & 0.120 & 0.304 & 0.290 & 0.020 & 0.012 & 0.020 \\ 
Llama-3.3-70B & 0.952 & 0.866 & 0.856 & 0.290 & 0.118 & 0.254 & 0.136 & 0.000 & 0.012 & 0.012 \\ 
Mixtral-8x7b & 0.466 & 0.282 & 0.108 & 0.002 & 0.032 & 0.024 & 0.056 & 0.000 & 0.000 & 0.022 \\ 
QwQ-32B-Preview & 0.912 & 0.504 & 0.498 & 0.164 & 0.058 & 0.106 & 0.124 & 0.000 & 0.002 & 0.020 \\ 
Llama3-8b-Instruct & 0.604 & 0.220 & 0.132 & 0.002 & 0.028 & 0.010 & 0.038 & 0.000 & 0.002 & 0.026 \\ 
\hline
+SFT & & & & & & & & & & \\ 
STaR(w/ GT) & 0.374 & 0.618 & 0.124 & 0.080 & 0.360 & 0.216 & 0.134 & 0.018 & 0.006 & 0.026 \\ 
STaR(w/ LLM) & 0.320 & 0.332 & 0.090 & 0.038 & 0.126 & 0.124 & 0.030 & 0.002 & 0.004 & 0.024 \\ 
\textbf{Llama-GT(w/o Thought)} & 0.804 & 0.864 & 0.340 & 0.302 & 0.652 & 0.652 & 0.370 & 0.024 & 0.020 & \textbf{0.068} \\ 
\textbf{Llama-GT} & 0.988 & \textbf{0.984} & 0.836 & \textbf{0.600} & \textbf{0.744} & \textbf{0.900} & \textbf{0.634} & \textbf{0.036} & \textbf{0.036} & 0.008 \\ 
\hline

\end{tabular}

}
\end{table*}

\clearpage
\section{Methods for constructing action thoughts and state thoughts for GCO problem}\label{app:a_s_method}

\subsection{Action thoughts generation methods}
For a GCO problem, there are mainly three classes of methods to construct action thought set $A$ with heuristics.  

\noindent - {\bf Ordinary Generation Rule}: Foundational strategies (e.g., greedy selection, random sampling) provide baseline mechanisms for generating thoughts. These rule-based approaches offer broad applicability across diverse problem domains through their simplicity.

\noindent - {\bf Simple Heuristic Thoughts:} Heuristics of GCO problems leverage structural properties of target problems to enhance operational efficiency. Such methods typically derive from simple reasoning conclusions of the problem.

\noindent - {\bf Complex Heuristic Thoughts:} Complex heuristics involve a broader set of operational primitives, presenting two fundamental challenges. First, their intricate implementation mechanisms lack intuitive interpretability. Second, the expanded solution space necessitates consideration of diverse operational combinations. These characteristics hinder LLMs from effectively discerning the underlying principles learning such heuristics.

\noindent - {\bf Mixed Heuristic Thoughts:} The former strategies possess distinct advantages, hybrid approaches demonstrate superior efficacy in specific problem contexts through strategic combination of complementary strategies.

\subsection{State thoughts generation methods}

In solving a GCO problem, maintaining instance state descriptions is essential for two reasons. First, the GCO instance evolves with each reasoning step, requiring the removal of redundant elements. Second, explicitly tracking instance states prevents hallucinations in LLMs. The following state thought generation methods are defined.
 
\noindent -  {\bf Instance Description:} Recording the current GCO instance information forms the foundational state representation. For example describe the node set $V$ and edge set $E$ of a graph $G$.

\noindent -  {\bf Instance Simplification:} Pruning redundant elements after every reasoning step. For example, in maximum independent set (MIS) problems, removing all neighbors of the current MIS set from the graph $G$.

\noindent -  {\bf Instance Reduction:} Transforming instances into equivalent problem spaces. For instance, solving MIS of $G$ can be reduced to finding minimum vertex cover (MVC) of $G$'s complement graph $\bar{G}$.

\noindent -  {\bf Solution Description:} The solution information not only emphasizes the importance of the solution but also shows the changes of the solution. This prompts the LLM to explicitly track how selected actions influence the current solution.

\noindent -  {\bf Solving flag:} Formal stopping criteria marking solution completion through state indicators.	

\section{Usual action thoughts and state thoughts for GCO problem}
\label{app:a_s}
 
In graph optimization problems, the fundamental heuristic operations primarily comprise four canonical primitives: node addition, node deletion, edge addition, and edge deletion. Formally, let $\mathbb{A}$ denote the action thought space for graph optimization, which includes the following important actions:
\begin{itemize}
    \item Add Nodes Based on the Given Optimal Solution ($a_1$): Add one or more nodes to the current solution set based on the given optimal solution.
    \item Add Nodes Based on Rules ($a_2$): Add one or more nodes to the current solution set using rules such as greedy or random selection.
    \item Add Nodes Based on Simple Prior Knowledge ($a_3$): Add one or more nodes to the current solution set using simple prior knowledge.
    \item Add Nodes Based on Complex Prior Knowledge ($a_4$): Add one or more nodes to the current solution set using complex prior knowledge.
    \item Remove Nodes Based on the Given Optimal Solution ($a_5$): Remove one or more nodes from the current solution set based on the given optimal solution.
    \item Remove Nodes Based on Rules ($a_6$): Remove one or more nodes from the current solution set using rules such as greedy or random selection.
    \item Remove Nodes Based on Simple Prior Knowledge ($a_7$): Remove one or more nodes from the current solution set using simple prior knowledge.
    \item Remove Nodes Based on Complex Prior Knowledge ($a_8$): Remove one or more nodes from the current solution set using complex prior knowledge.
    \item Add Edges Based on the Given Optimal Solution ($a_9$): Add one or more edges to the current solution set based on the given optimal solution.
    \item Add Edges Based on Rules ($a_{10}$): Add one or more edges to the current solution set using rules such as greedy or random selection.
    \item Add Edges Based on Simple Prior Knowledge ($a_{11}$): Add one or more edges to the current solution set using simple prior knowledge.
    \item Add Edges Based on Complex Prior Knowledge ($a_{12}$): Add one or more edges to the current solution set using complex prior knowledge.
    \item Remove Edges Based on the Given Optimal Solution ($a_{13}$): Remove one or more edges from the current solution set based on the given optimal solution.
    \item Remove Edges Based on Rules ($a_{14}$): Remove one or more edges from the current solution set using rules such as greedy or random selection.
    \item Remove Edges Based on Simple Prior Knowledge ($a_{15}$): Remove one or more edges from the current solution set using simple prior knowledge.
    \item Remove Edges Based on Complex Prior Knowledge ($a_{16}$): Remove one or more edges from the current solution set using complex prior knowledge.
\end{itemize}

The fundamental state thought space $\mathbb{S}$ for graph optimization problems is defined through node and edge set representations. Formally, the main canonical state components are structured as follows:

\begin{itemize}
    \item Solving State ($s_1$): Describes the solving state to determine whether the process is complete.
    \item Add Nodes to the Graph ($s_2$): Describes the operation of adding nodes to the original graph for subsequent solving.
    \item Remove Nodes from the Graph ($s_3$): Describes the operation of removing nodes from the original graph for subsequent solving.
    \item Add Edges to the Graph ($s_4$): Describes the operation of adding edges to the original graph for subsequent solving.
    \item Remove Edges from the Graph ($s_5$): Describes the operation of removing edges from the original graph for subsequent solving.
    \item Graph Node Set ($s_6$): Describes the set of all nodes or partial nodes in the original graph.
    \item Graph Edge Set ($s_7$): Describes the set of all edges or partial edges in the original graph.
    \item Current Solution Set ($s_8$): Describes the set of elements in the current solution.
    \item Final Solution Set ($s_9$): Describes the set of elements in the final solution.
\end{itemize}


\section{Concrete Applications of Forward and Backward MTP Frameworks}

\subsection{A Forward MTP for the Connected Components Problem}\label{sec:forward_c_c}

\begin{algorithm}[htbp]
\caption{ A Thoughts Template for Connected Component Problem }\label{algo:template_cc}
\begin{algorithmic}[1]
    \REQUIRE Graph $G = (V,E)$
    \STATE $CC \leftarrow \emptyset$;\COMMENT{Initialize connected components}
    \WHILE{$V \ne \emptyset$}
        \STATE Action: choose a node for BFS randomly;
        \STATE $u \leftarrow \text{RandomSelect}(G)$; \COMMENT{Random node selection}
        \STATE $C_u \leftarrow \text{BFS}(G, u)$;\COMMENT{Component discovery}
        \STATE $CC \leftarrow CC \cup \{C_u\}$;
        \STATE $V \leftarrow V \setminus C_u$;\COMMENT{Graph simplification}
    \ENDWHILE
    \STATE State: describe connected components in $CC$;
    \STATE {\bf return} $CC$;
\end{algorithmic}
\end{algorithm}

\begin{algorithm}[htbp]
\caption{ Breadth-First Search($G$,$u$)}\label{algo:template_bfs}
\begin{algorithmic}[1]
    \REQUIRE Graph $G$, start node $u$
    \ENSURE Connected component $C$
    \STATE Action: Start BFS at node $u$ of $G$;
    \STATE $L \leftarrow \{u\}$; \COMMENT{ Nodes waiting to be visited}
    \STATE $C \leftarrow \emptyset$; \COMMENT{ Visited nodes }
    \WHILE{$L \ne \emptyset$}
        \STATE $v \leftarrow$ PopFrom(L);\COMMENT{Select unvisited node}
        \STATE Action: add $v$ to the current component;
        \STATE $C \leftarrow C \cup \{v\}$;
        \FOR{$w \in$ Neighbor($G$,$v$)}
            \IF{$w \notin C \land w\notin L$} 
                \STATE \text{AddTo}($w, L$);\COMMENT{Record unvisited nodes}
                \STATE State: add an unvisited neighbor $w$ to $L$;
            \ENDIF
        \ENDFOR
        \STATE State: show the current visited nodes $C$;
    \ENDWHILE
    \STATE State: finished, show the connected component $C$;
    \STATE {\bf return} $C$;
\end{algorithmic}
\end{algorithm}

The Connected Components (CC) problem requires identifying all maximally connected subgraphs in an undirected graph $G$. Formally, given $G=(V,E)$, the goal is to partition $V$ into disjoint subsets $\{C_1,\ldots,C_k\}$ where each $C_i$ forms a connected subgraph.

{\bf Action Thought Generation Methods:}
\begin{itemize}
    \item {\bf A simple heuristic thought: Breadth-First Search (BFS)}, systematically explores node neighborhoods through queue-based traversal (lines 1 and 6 of Algorithm~\ref{algo:template_bfs}).
    \item {\bf An ordinary generation rule: Random Selection}, chooses initial nodes for BFS through random sampling (line 3 of Algorithm \ref{algo:template_cc} ).
\end{itemize}
These action thoughts in the three lines constitute the set $A$.

{\bf State Thought Generation Methods:}
\begin{itemize}
    \item {\bf Instance Simplification}: Add an unvisited node to the queue of nodes to be visited (line 11 of Algorithm~\ref{algo:template_bfs}).
    \item {\bf Solution Description}: Current connected component being explored (line 14 of Algorithm~\ref{algo:template_bfs},  line 9 of Algorithm \ref{algo:template_cc}).
    \item {\bf Solving Flag}: Indicator for algorithm completion (line 16 of Algorithm~\ref{algo:template_bfs} ).
\end{itemize}
These state thoughts in the four lines constitute the set $S$.

A forward MTP for CC is shown in Algorithm \ref{algo:template_cc}. $\text{RandomSelect}(G)$ select randomly a node $u$ of $G$.

In BFS, the queue $L$ stores nodes awaiting visitation, while set $C$ maintains all visited nodes. When $L \neq \emptyset$, the algorithm selects a node $v$ of $L$ to visit, then adds $v$'s unvisited neighbors to $L$. PopFrom(L) pops the first node of $L$. Neighbor($G$,$v$) returns all neighbor node of $v$ in $G$. AddTo($w$, $L$) appends $w$ to the end of $L$.

\subsection{A Backward MTP for the MIS Problem}\label{sec:backward_mis}

\begin{algorithm}[htbp]
\caption{ A Thought Template for MIS}\label{algo:template_mis}
\begin{algorithmic}[1]
\REQUIRE Graph $G = (V, E)$, a mis solver $\mathcal{X}$.
\STATE $OPT\leftarrow \mathcal{X}(G)$;\COMMENT{Compute optimal MIS}
\STATE $MIS \leftarrow \emptyset$;\COMMENT{Initialize solution}
\WHILE{$V \ne \emptyset$}
    \STATE Action: add all isolated nodes to $MIS$;
    \STATE $MIS \leftarrow MIS \cup \text{Isolated}(G)$;\COMMENT{Add all isolated nodes}
    \STATE Action: add one node of $OPT$ to  $MIS$;
    \STATE $u \leftarrow \text{AddOne}(OPT)$;\COMMENT{Add optimal node}
    \STATE $MIS \leftarrow MIS \cup\{ u\}$;
    \STATE State: describe the current $MIS$;
    \STATE State: delete all neighbors of $u$ in G;
    \STATE $V \leftarrow V \setminus (\text{Isolated}(G) \cup \text{Neighbor}(G,u)  \cup \{u\})$; \COMMENT{ Remove useless nodes}
    \STATE $OPT \leftarrow OPT \setminus MIS$;
    \STATE State: describe the current $G$;
\ENDWHILE
\STATE State: finished, describe the solution $MIS$;
\STATE {\bf return} $MIS$.
\end{algorithmic}
\end{algorithm}
    
The MIS problem identifies the largest subset of non-adjacent nodes. The AddOne thought comprises two specialized operations:
\begin{itemize}
    \item \textbf{AddOne: Add Isolated Nodes}, immediately incorporates all isolated nodes into the current solution $MIS$ (line 4 of Algorithm \ref{algo:template_mis}).
    \item \textbf{AddOne: Add Optimal Nodes}, selectively integrates one node from solver outputs into the current solution $MIS$ (line 6 of Algorithm \ref{algo:template_mis}).
\end{itemize}
The first action atomically adds all isolated nodes to the current solution because all isolated nodes belong to the optimal MIS solution. These action thoughts in the two lines constitute the set $A$.

{\bf State Thought Generation Methods:}
\begin{itemize}
    \item {\bf Instance Description}: Maintains graph $G$'s current structure (line~13 of Algorithm \ref{algo:template_mis})
    \item {\bf Instance Simplification}: Records graph simplification operations (line~10 of Algorithm \ref{algo:template_mis}) 
    \item {\bf Solution Description}: Tracks current/final solution candidates (lines~9 \&~15 of Algorithm \ref{algo:template_mis})
    \item {\bf Solving Flag}: Monitors termination conditions (line 15 of Algorithm \ref{algo:template_mis})
\end{itemize}
These state thoughts in the four lines constitute the set $S$.

A backward MTP for MIS  is shown as follows. An integer programming model of MIS problem is put into Gurobi, which serves as an optimal solution solver. Isolated($G$) returns all nodes of $G$ without neighbors.

\section{Ablation and Supplementary Experiments}
\label{appendix:ablation}

\subsection{Impact of Thought Types: State vs Action}

We conducted a focused ablation to understand the role of state vs. action thoughts. The results show that use only action thoughts yields worse performance than even the no-thought baseline, suggesting that action thoughts alone lead to unstructured, less interpretable reasoning chains. The integration of state thoughts plays a crucial role in structuring the reasoning process, enabling more effective optimization.

\begin{table}[h]
\centering
\caption{Ablation: Effect of Thought Types}
\begin{tabular}{lcc}
\toprule
Method & Polynomial Task & NP-Hard Task \\
\midrule
Llama3-8B (w/o thoughts)     & 0.752 & 0.509 \\
Llama3-8B (action-only)      & 0.741 & 0.488 \\
Llama3-8B (full thoughts)    & \textbf{0.920} & \textbf{0.552} \\
\bottomrule
\end{tabular}
\label{tab:thought_ablation}
\end{table}

\subsection{Comparison with Classical Heuristics on MIS}

We benchmarked Llama-GT against classical baselines including Greedy, Luby's Algorithm\cite{luby_1986}, and BDOne\cite{chang2017computing}. As the Best-of-N (BoN) value increases, Llama-GT demonstrates significant performance improvements. On both easy and hard instances, Llama-GT with higher BoN values not only surpasses heuristic approaches but also achieves performance comparable to the Gurobi solver, particularly excelling in high-difficulty settings.

\begin{table}[h]
\centering
\caption{Comparison on MIS Problem (Average Solution Size)}
\begin{tabular}{lcc}
\toprule
Method & MIS\_easy & MIS\_hard \\
\midrule
Llama-GT (N=1)   & 4.472 & 11.844 \\
Llama-GT (N=16)  & 4.502 & 12.568 \\
Llama-GT (N=32)  & 4.502 & \textcolor{red}{\textbf{12.570}} \\
Greedy           & 4.502 & 12.562 \\
Luby             & 4.502 & 12.568 \\
BDOne            & 4.502 & 12.568 \\
Gurobi           & 4.502 & \textcolor{red}{\textbf{12.570}} \\
\bottomrule
\end{tabular}
\label{tab:mis_comparison}
\end{table}

\subsection{Relative Improvements Across Different Model Scales}
\begin{table}[h]
\centering
\caption{Performance Across Model Scales, *inf means the pre-finetuning result is 0.0}
\begin{tabular}{lcc}
\toprule
\textbf{Task} & \textbf{Llama3.2-3B\footnote{\url{https://huggingface.co/meta-llama/Llama-3.2-3B-Instruct}}} & \textbf{Llama3-8B} \\
\midrule
MIS & 0.988 (+766.7\%) & 0.994 (+300.8\%) \\
Diameter & 0.872 (+2625.0\%) & 0.954 (+1262.9\%) \\
MCS & 0.394 (+310.4\%) & 0.496 (+123.4\%) \\
MIS (Hard) & 0.870 (+21650.0\%) & 0.900 (+8900.0\%) \\
Diameter (Hard) & 0.358 (inf) & 0.600 (+29900.0\%) \\
MCS (Hard) & 0.018 (inf) & 0.036 (+1700.0\%) \\
\bottomrule
\end{tabular}
\label{tab:model_scales}
\end{table}

Due to computational constraints, we could not fine-tune the Llama3-70B model, but instead evaluated generalization trends using smaller-scale models. Our analysis reveals that the relative performance gain from thought integration diminishes with increasing model size, suggesting that larger models inherently possess stronger reasoning capabilities. Nevertheless, thought guidance remains valuable—it significantly enhances the capabilities of smaller models, potentially providing a more computationally efficient alternative to pure model scaling. We recommend that future research investigate this hypothesis across a broader range of model sizes to better understand the scaling dynamics.

\section{Performance Degradation on Certain NP-hard Problems}
\label{app:performace_degradation}
While our method demonstrates strong performance on specific NP-hard tasks such as MIS, achieving significant improvements over baseline models, we observe relatively modest gains for tasks like TSP, MCS, and GED. In certain instances, performance metrics even regressed below those of baseline models trained without thought mechanisms, particularly for the GED task.

We posit two principal factors contributing to this performance disparity:

First, these tasks inherently exhibit \textbf{higher computational complexity} compared to problems like MIS or MCP, presenting greater challenges for LLM-based solutions. 

Second, the thought construction process for TSP, MCS, and GED primarily utilizes a single action type (e.g., ``Add Nodes Based on the Given Optimal Solution,'' detailed in Appendix~\ref{app:a_s}). This monolithic approach operates as a \textbf{black-box mechanism}, limiting the incorporation of domain-specific prior knowledge or structural knowledge. Conversely, tasks like MIS benefit from diverse action types (e.g., both ``Add Nodes Based on the Given Optimal Solution'' and ``Add Nodes Based on Simple Prior Knowledge'') and state transformations (e.g., ``Remove Nodes from the Graph''). These mechanisms effectively \textbf{embed structural priors} (e.g., mandatory inclusion of isolated nodes in independent sets) and \textbf{systematically prune the search space}, thereby reducing hallucination risks while enhancing solution quality. For detailed thought construction examples, please refer to Appendix~\ref{app:thought_case}.

\section{Automated Thought Dataset Synthesis via LLM-Driven Code Generation}
\label{app:llm_gen}
We created a template to automate dataset generation for ten tasks using a LLM(Qwen2.5-Coder-32B-Instruct\footnote{\url{https://huggingface.co/Qwen/Qwen2.5-Coder-32B-Instruct}}). 
This automated approach generated valid code for the tasks, creating datasets of equivalent size to fine-tune the Meta-Llama-3-8B-Instruct model. 
We compared four configurations: 
1) Origin: The base model, 
2) w/o Thought: Fine-tuned with direct-answer supervision, 
3) Llama-GT (w/ LLM-Design): Fine-tuned on datasets generated by LLM-synthesized code, and 
4) Llama-GT (w/ Human-Design): Fine-tuned on datasets generated by human-designed code.

\begin{figure}[ht]
\begin{center}
\centerline{\includegraphics[width=\columnwidth]{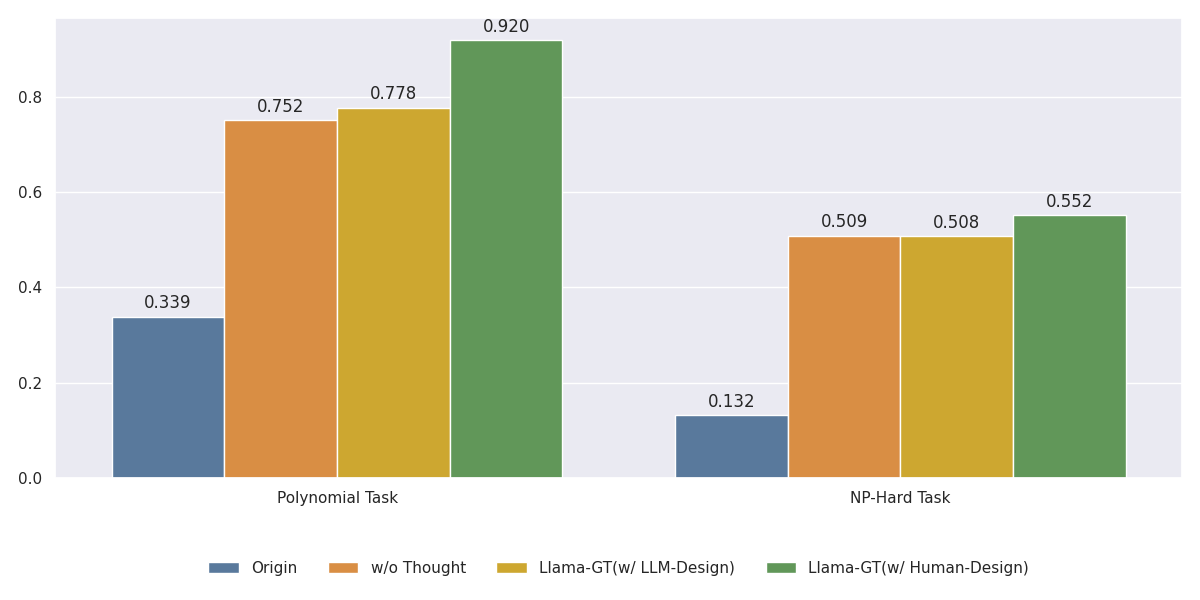}}
\caption{Performance comparison of Meta-Llama-3-8B-Instruct variants on Polynomial and NP-hard tasks: Base model (Origin), fine-tuned with direct-answer datasets (w/o Thought),datasets generated using LLM-synthesized code Llama-GT (w/LLM-Design), and datasets generated using human-designed code Llama-GT (w/Human-Design). The values represent the average ratio of the number of optimal solutions achieved across different tasks.}
\label{llm-generated}
\end{center}
\end{figure}

As shown in Figure \ref{llm-generated}, Llama-GT (w/ LLM-Design) significantly improves reasoning capabilities over the Origin model, outperforming w/o Thought on Polynomial tasks. 
However, on NP-hard tasks, the LLM-generated code struggles to produce high-quality thought sequences, resulting in performance similar to w/o Thought ($0.509$ vs. $0.508$). 
While Llama-GT (w/LLM-Design) lags behind Llama-GT (w/Human-Design) in both categories, it offers advantages like lower construction costs for unseen problems and better compatibility with optimization methods. 
Future work will focus on developing frameworks to improve the quality and generality of LLM-generated datasets.

\clearpage

\subsection{Automated Thought Dataset Generation Template}
\label{app:prompt}

\subsubsection{Outer Prompt}
\begin{lstlisting}

You are a professional mathematician and computer scientist. I am working on solving a graph theory problem called {TASK NAME}. {TASK DESCRIPTON} I want to reconstruct the steps involved in solving this graph theory problem starting from the optimal solution. Please help me construct a subset of combinations of states and actions from the following state thought space and action thought space, so that the algorithm using this subset can clearly and step-by-step output the optimal reasoning process for solving {task.task_full_name} in textual form. Please do not use any third-party libraries or known algorithms to simply obtain the final answer. What we need is the reasoning process for solving the problem in as much detail as possible.
The states describe the transformations and intermediate stages of the graph during the solution process. Each element in the state thought space is defined and explained as follows:
1. Solving State: Describes the solving state to determine whether the process is complete.
2. Add Nodes to the Graph: Describes the operation of adding nodes to the original graph for subsequent solving.
3. Remove Nodes from the Graph: Describes the operation of removing nodes from the original graph for subsequent solving.
4. Add Edges to the Graph: Describes the operation of adding edges to the original graph for subsequent solving.
5. Remove Edges from the Graph: Describes the operation of removing edges from the original graph for subsequent solving.
6. Graph Node Set: Describes the set of all nodes or partial nodes in the original graph.
7. Graph Edge Set: Describes the set of all edges or partial edges in the original graph.
8. Current Solution Set: Describes the set of elements in the current solution.
9. Final Solution Set: Describes the set of elements in the final solution.
The actions represent operations performed on the solution set. Each element in the action thought space is defined and explained as follows:
1. Add Nodes Based on the Given Optimal Solution: Add one or more nodes to the current solution set based on the given optimal solution.
2. Add Nodes Based on Rules: Add one or more nodes to the current solution set using rules such as greedy or random selection.
3. Add Nodes Based on Simple Prior Knowledge: Add one or more nodes to the current solution set using simple prior knowledge.
4. Add Nodes Based on Complex Prior Knowledge: Add one or more nodes to the current solution set using complex prior knowledge.
5. Remove Nodes Based on the Given Optimal Solution: Remove one or more nodes from the current solution set based on the given optimal solution.
6. Remove Nodes Based on Rules: Remove one or more nodes from the current solution set using rules such as greedy or random selection.
7. Remove Nodes Based on Simple Prior Knowledge: Remove one or more nodes from the current solution set using simple prior knowledge.
8. Remove Nodes Based on Complex Prior Knowledge: Remove one or more nodes from the current solution set using complex prior knowledge.
9. Add Edges Based on the Given Optimal Solution: Add one or more edges to the current solution set based on the given optimal solution.
10. Add Edges Based on Rules: Add one or more edges to the current solution set using rules such as greedy or random selection.
11. Add Edges Based on Simple Prior Knowledge: Add one or more edges to the current solution set using simple prior knowledge.
12. Add Edges Based on Complex Prior Knowledge: Add one or more edges to the current solution set using complex prior knowledge.
13. Remove Edges Based on the Given Optimal Solution: Remove one or more edges from the current solution set based on the given optimal solution.
14. Remove Edges Based on Rules: Remove one or more edges from the current solution set using rules such as greedy or random selection.
15. Remove Edges Based on Simple Prior Knowledge: Remove one or more edges from the current solution set using simple prior knowledge.
16. Remove Edges Based on Complex Prior Knowledge: Remove one or more edges from the current solution set using complex prior knowledge.
Here, I'll give you two examples about what to output:
Example 1:
{
    "input": "In an undirected graph, (i,j) means that node i and node j are connected with an undirected edge. I'll give an instance of a graph, please help me find the maximum independent set in the graph and list the steps and results for each iteration. The Maximum Independent Set problem is an optimization problem in graph theory that aims to identify the largest set of vertices in a graph, where no two vertices in the set are adjacent."
    "output": {"Solving State":  "Determine whether the current graph is an empty graph.", "Add Nodes Based on Simple Prior Knowledge": "Add isolated nodes: [].", "Add Nodes Based on the Given Optimal Solution": "Add the most appropriate node: [].", "Remove Nodes from the Graph": "Remove the neighboring nodes of the node: [].", "Graph Node Set": "The remaining nodes of the graph are: [].", "Current Solution Set": "The current Independent Set is: [].", "Final Solution Set": "The maximum independent set is []."}
}
Example 2:
{
    "input": "In an undirected graph, (i,j) means that node i and node j are connected with an undirected edge. I'll give an instance of a graph and two nodes, please help me find the common neighbor nodes of the given two nodes in the graph."
    "output": {"Graph Node Set": "The neighboring nodes of the node <chosen node 1>: [].", "Graph Node Set": "The neighboring nodes of the node <chosen node 2>: [].", "Final Solution Set": "The common neighbor nodes of the two nodes are: []."}
}
Please provide the state and action combinations along with their descriptions for {TASK NAME} based on the format of "output" above. {TASK DESCRIPTION} Do not provide any explanations or descriptions related to "output" and there is no need to provide any examples. Since the nodes and edges of the original graph will be provided in the input, in order to keep the output inference text brief, please avoid choosing Graph Node Set and Graph Edge Set to describe the original graph unless necessary. 

\end{lstlisting}

\subsubsection{Inner Prompt}

\begin{lstlisting}

You are a professional mathematician and computer science expert. I am working on solving a graph theory problem called {task.task_full_name}. {task.task_description()} I would like to reconstruct the steps involved in solving this graph theory problem starting from the optimal solution. Please help me build a Python function using all the provided combinations of states and actions, so that this function can return the reasoning steps to solve the problem. Please do not use any third-party libraries or known algorithms to simply obtain the final answer. What we need is the reasoning process for solving the problem in as much detail as possible.
Here is two examples:
Example 1:
{
	"input": "In an undirected graph, (i,j) means that node i and node j are connected with an undirected edge. I'll give an instance of a graph, please help me find the maximum independent set in the graph and list the steps and results for each iteration. The Maximum Independent Set problem is an optimization problem in graph theory that aims to identify the largest set of vertices in a graph, where no two vertices in the set are adjacent. Here's the combination of states and actions constructed from expert: {{"Solving State":  "Determine whether the current graph is an empty graph.", "Add Nodes Based on Simple Prior Knowledge": "Add isolated nodes: [].", "Add Nodes Based on the Given Optimal Solution": "Add the most appropriate node: [].", "Remove Nodes from the Graph": "Remove the neighboring nodes of the node: [].", "Graph Node Set": "The remaining nodes of the graph are: [].", "Current Solution Set": "The current Independent Set is: [].", "Final Solution Set": "The maximum independent set is []."}}"
    "output": ```python
def mis_optimal_trace(G_: nx.Graph, optimal_solution: List[int]) -> Tuple[str, List[int]]:
    G = G_.copy()
    s = ""
    mis = optimal_solution.copy()
    I = []
    while G: # Solving State
        Isopnts = [u for u in G if len(list(G.neighbors(u))) == 0]
        I.extend(Isopnts)
        s += f"Add isolated nodes: {{list(set(Isopnts))}}.\\n" # Add Nodes Based on Simple Prior Knowledge
        G.remove_nodes_from(Isopnts)
        if not G:
            s += f"The remaining nodes of the graph are: {{list(G.nodes)}}.\\n" # Graph Node Set
            break
        
        mis = [elem for elem in mis if elem not in I]
        chosen_node = random.choice(mis)
        I.append(chosen_node)
        s += f"Add the most appropriate node: {{chosen_node}}.\\n" # Add Nodes Based on the Given Optimal Solution
        s += f"Remove the neighboring nodes of the node {{chosen_node}}: {{list(G.neighbors(chosen_node))}}.\\n" # Remove Nodes from the Graph

        G.remove_nodes_from(list(G.neighbors(chosen_node)))
        G.remove_node(chosen_node)
        s += f"The remaining nodes of the graph are: {{list(G.nodes)}}.\\n" # Graph Node Set
        s += f"The current Independent Set is: {{I}}.\\n" # Current Solution Set
    
    s += "Finished!\\n"
    s += f'The maximum independent set is {{mis}}.' # Final Solution Set
    return s, I
```
}
Example 2:
{
	"input": "In an undirected graph, (i,j) means that node i and node j are connected with an undirected edge. I'll give an instance of a graph and two nodes, please help me find the common neighbor nodes of the given two nodes in the graph. Here's the combination of states and actions constructed from expert: {{"Graph Node Set": "The neighboring nodes of the node <chosen node 1>: [].", "Graph Node Set": "The neighboring nodes of the node <chosen node 2>: [].", "Final Solution Set": "The common neighbor nodes of the two nodes are: []."}}"
    "output": ```python
def neighbor_optimal_trace(G_: nx.Graph, chosen_nodes: List[int], optimal_solution: List[int]) -> Tuple[str, List[int]]:
    G = G_.copy()
    s = ""
    neighbor_node_1 = list(G.neighbors(chosen_nodes[0]))
    neighbor_node_2 = list(G.neighbors(chosen_nodes[1]))
    common_neighbor_nodes = list(set(neighbor_node_1) & set(neighbor_node_2))
    s += f"The neighboring nodes of the node {{chosen_nodes[0]}}: {{neighbor_node_1}}.\\n" # Graph Node Set
    s += f"The neighboring nodes of the node {{chosen_nodes[1]}}: {{neighbor_node_2}}.\\n" # Graph Node Set
    s += f"The common neighbor nodes of the two nodes are: {{common_neighbor_nodes}}.\\n" # Final Solution Set
    return s, common_neighbor_nodes
```
}
Here's the input: "{TASK DESCRIPTION} Here's the combination of states and actions constructed from expert: {OUTER RESPONSE}."
Please provide the python code of solution for {TASK NAME} based on the format of "output" above. Do not provide any explanations or descriptions related to "output," and there is no need to provide any examples. Please use all the provided states and actions set to construct the function. Here's the function definition header: "{TASK FUNCTION HEADER}". Please construct the optimal Python code to solve the problem. Ensure the code is correct, the reasoning steps are complete, and the code is highly readable, while minimizing the length of the reasoning text as much as possible. And you must concatenate '{TASK OUTPUT FORMAT}' at the end of your code, where <final answer> is a python list. Do not modify, add, or delete any words or symbols in the statement, and do not change case.

\end{lstlisting}

\subsubsection{Example}

Here, we use the Minimum Vertex Cover problem as an example to demonstrate the Outer output and Inner output generated using the above-mentioned template. The problem description is as follows.

\begin{lstlisting}
In an undirected graph, (i,j) means that node i and node j are connected with an undirected edge. I'll give an instance of a graph, please help me find the minimum vertex cover in the graph and list the steps and results for each iteration. The Minimum Vertex Cover problem is an optimization problem in graph theory that aims to find the smallest set of vertices in a graph such that every edge in the graph is incident to at least one vertex in the set.
\end{lstlisting}

{\bf{Outer Output}}

\begin{lstlisting}

{'Solving State': 'Determine whether the current graph is an empty graph.', 'Add Nodes Based on Simple Prior Knowledge': 'Add nodes with the highest degree: [].', 'Add Nodes Based on the Given Optimal Solution': 'Add the most appropriate node: [].', 'Remove Nodes from the Graph': 'Remove the selected node and its incident edges: [].', 'Graph Node Set': 'The remaining nodes of the graph are: [].', 'Graph Edge Set': 'The remaining edges of the graph are: [].', 'Current Solution Set': 'The current Vertex Cover is: [].', 'Final Solution Set': 'The minimum vertex cover is [].'}

\end{lstlisting}

{\bf{Inner Output}}

\begin{lstlisting}

def mvc_optimal_trace(G_: nx.Graph, optimal_solution: List[int]) -> str:
    G = G_.copy()
    s = ""
    mvc = optimal_solution.copy()
    C = []
    while G:  # Solving State
        if not G.edges:
            s += "The graph is empty.\n"
            break
        
        # Add Nodes Based on Simple Prior Knowledge
        degrees = dict(G.degree())
        max_degree_node = max(degrees, key=degrees.get)
        C.append(max_degree_node)
        s += f"Add nodes with the highest degree: {max_degree_node}.\n"
        
        # Remove Nodes from the Graph
        neighbors = list(G.neighbors(max_degree_node))
        G.remove_node(max_degree_node)
        s += f"Remove the selected node and its incident edges: {max_degree_node}.\n"
        
        # Graph Node Set and Edge Set
        s += f"The remaining nodes of the graph are: {list(G.nodes)}.\n"
        s += f"The remaining edges of the graph are: {list(G.edges)}.\n"
        
        # Current Solution Set
        s += f"The current Vertex Cover is: {C}.\n"
    
    s += "Finished!\n"
    s += f'The minimum vertex cover is {mvc}.'  # Final Solution Set
    return s
\end{lstlisting}

\section{Detailed Case Studies of Thought Processes}
\label{app:thought_case}
In this section, we present ten concrete task examples accompanied by their corresponding reasoning thoughts to systematically demonstrate the thought construction methodology.

\begin{table*}[h]
\centering
\small
\caption{An example of the Connected Component task.} 
\begin{tabular}{p{0.95\linewidth}}
\toprule

\textbf{Task Description}

In an undirected graph, (i,j) means that node i and node j are connected with an undirected edge. I'll give an instance of a graph, and the task is to identify one representative node from each connected component within the graph. \\\\

\textbf{Input}

The graph has 7 nodes. The nodes are numbered from 0 to 6, and the edges are: [(0, 3), (2, 6), (3, 4), (4, 6)]. \\\\

\textbf{Output}

Choose node 0 as the start point of the current connectivity component.\\
Add node 0 into the connected component list.\\
Add the unvisited neighboring nodes of the node 0 into the search queue: [3].\\
Add node 3 into the connected component list.\\
Add the unvisited neighboring nodes of the node 3 into the search queue: [4].\\
Add node 4 into the connected component list.\\
Add the unvisited neighboring nodes of the node 4 into the search queue: [6].\\
The current connected component is: [0, 3, 4].\\
Add node 6 into the connected component list.\\
Add the unvisited neighboring nodes of the node 6 into the search queue: [2].\\
Add node 2 into the connected component list.\\
Add the unvisited neighboring nodes of the node 2 into the search queue: [].\\
Finished!\\
Connected component 1: Nodes = [0, 3, 4, 6, 2], Representative node = 0.\\
Choose node 1 as the start point of the current connectivity component.\\
Add node 1 into the connected component list.\\
Add the unvisited neighboring nodes of the node 1 into the search queue: [].\\
Finished!\\
Connected component 2: Nodes = [1], Representative node = 1.\\
Choose node 5 as the start point of the current connectivity component.\\
Add node 5 into the connected component list.\\
Add the unvisited neighboring nodes of the node 5 into the search queue: [].\\
Finished!\\
Connected component 3: Nodes = [5], Representative node = 5.\\
The representative nodes for each connected component are: [0, 1, 5]. \\

\bottomrule

\label{table:Connected Component}
\end{tabular}

\vspace{-5mm}
\end{table*}

\begin{table*}[h]
\centering
\small
\caption{An example of the Shortest Distance task.} 
\begin{tabular}{p{0.95\linewidth}}
\toprule

\textbf{Task Description}

In an undirected graph, (i,j) means that node i and node j are connected with an undirected edge. I'll give an instance of a graph, please help me find the shortest path between two nodes in the graph and list the steps and results for each iteration. The shortest path between two nodes is defined as the path with the fewest edges that connect the source node to the target node. If no path exists, output an empty path. \\\\

\textbf{Input}

The graph has 9 nodes. The nodes are numbered from 0 to 8, and the edges are: [(0, 7), (0, 8), (1, 3), (1, 8), (2, 6), (3, 6), (4, 8), (6, 7), (7, 8)]. The source node is 4, and the target node is 3. \\\\

\textbf{Output}

Current path: [4], the neighboring nodes of the node 4: [8].\\
Current path: [4, 8], the neighboring nodes of the node 8: [0, 1, 4, 7].\\
Current path: [4, 8, 0], the neighboring nodes of the node 0: [7, 8].\\
Current path: [4, 8, 1], the neighboring nodes of the node 1: [3, 8].\\
Found the target node 3.\\
The shortest path is [4, 8, 1, 3]. \\

\bottomrule

\label{table:Shortest Distance}
\end{tabular}

\vspace{-5mm}
\end{table*}

\begin{table*}[h]
\centering
\small
\caption{An example of the Common Neighbor task.} 
\begin{tabular}{p{0.95\linewidth}}
\toprule

\textbf{Task Description}

In an undirected graph, (i,j) means that node i and node j are connected with an undirected edge. I'll give an instance of a graph and two nodes, please help me find the common neighbor nodes of the given two nodes in the graph. \\\\

\textbf{Input}

The graph has 10 nodes. The nodes are numbered from 0 to 9, and the edges are: [(0, 1), (0, 3), (0, 5), (0, 7), (0, 8), (1, 3), (1, 6), (1, 7), (1, 8), (1, 9), (2, 3), (2, 4), (2, 7), (2, 8), (2, 9), (3, 5), (3, 6), (3, 7), (3, 9), (4, 7), (4, 9), (5, 6), (5, 8), (6, 8), (6, 9), (7, 8), (7, 9), (8, 9)]. The given nodes are [7, 3]. \\\\

\textbf{Output}

The neighboring nodes of the node 7: [0, 1, 2, 3, 4, 8, 9].\\
The neighboring nodes of the node 3: [0, 1, 2, 5, 6, 7, 9].\\
The common neighbor nodes of the two nodes are: [0, 1, 2, 9]. \\

\bottomrule

\label{table:Common Neighbor}
\end{tabular}

\vspace{-5mm}
\end{table*}

\begin{table*}[h]
\centering
\small
\caption{An example of the Minimum Vertex Cover task.} 
\begin{tabular}{p{0.95\linewidth}}
\toprule

\textbf{Task Description}

In an undirected graph, (i,j) means that node i and node j are connected with an undirected edge. I'll give an instance of a graph, please help me find the minimum vertex cover in the graph and list the steps and results for each iteration. The Minimum Vertex Cover problem is an optimization problem in graph theory that aims to find the smallest set of vertices in a graph such that every edge in the graph is incident to at least one vertex in the set. \\\\

\textbf{Input}

The graph has 10 nodes. The nodes are numbered from 0 to 9, and the edges are: [(1, 5), (1, 6), (2, 7)]. \\\\

\textbf{Output}

Remove isolated nodes: [0, 3, 4, 8, 9].\\
Add the most appropriate node: 1.\\
The current Vertex Cover is: [1].\\
Remove the edges of node 1: [(1, 5), (1, 6)].\\
Remove isolated nodes: [1, 5, 6].\\
Add the most appropriate node: 2.\\
The current Vertex Cover is: [1, 2].\\
Remove the edges of node 2: [(2, 7)].\\
Remove isolated nodes: [2, 7].\\
There is no edge left in the graph.\\
Finished!\\
The minimum vertex cover is [1, 2]. \\

\bottomrule

\label{table:Minimum Vertex Cover}
\end{tabular}

\vspace{-5mm}
\end{table*}

\begin{table*}[h]
\centering
\small
\caption{An example of the Maximum Independent Set task.} 
\begin{tabular}{p{0.95\linewidth}}
\toprule

\textbf{Task Description}

In an undirected graph, (i,j) means that node i and node j are connected with an undirected edge. I'll give an instance of a graph, please help me find the maximum independent set in the graph and list the steps and results for each iteration. The Maximum Independent Set problem is an optimization problem in graph theory that aims to identify the largest set of vertices in a graph, where no two vertices in the set are adjacent. \\\\

\textbf{Input}

The graph has 8 nodes. The nodes are numbered from 0 to 7, and the edges are: [(3, 4), (5, 6)]. \\\\

\textbf{Output}

Add isolated nodes: [0, 1, 2, 7].\\
Add the most appropriate node: 3.\\
The current Independent Set is: [0, 1, 2, 7, 3].\\
Remove the neighboring nodes of the node 3: [4].\\
The remaining nodes of the graph are: [5, 6].\\
Add isolated nodes: [].\\
Add the most appropriate node: 5.\\
The current Independent Set is: [0, 1, 2, 7, 3, 5].\\
Remove the neighboring nodes of the node 5: [6].\\
The remaining nodes of the graph are: [].\\
Finished!\\
The maximum independent set is [0, 1, 2, 7, 3, 5]. \\

\bottomrule

\label{table:Maximum Independent Set}
\end{tabular}

\vspace{-5mm}
\end{table*}

\begin{table*}[h]
\centering
\small
\caption{An example of the Graph Diameter task.} 
\begin{tabular}{p{0.95\linewidth}}
\toprule

\textbf{Task Description}

In an undirected graph, (i,j) means that node i and node j are connected with an undirected edge. I'll give an instance of a graph, please help me find the diameter of the graph and list the steps and results for each iteration. The diameter of a graph is the longest shortest path between any two nodes in the graph. \\\\

\textbf{Input}

The graph has 10 nodes. The nodes are numbered from 0 to 9, and the edges are: [(0, 1), (0, 9), (1, 3), (2, 3), (3, 4), (3, 6), (4, 6), (5, 9), (6, 8), (7, 8)]. \\\\

\textbf{Output}

Choose the most appropriate node as source node of the diameter path: 7.\\
Calculating the longest path among all the shortest paths from the graph and source node 7.\\
The neighboring nodes of the node 7: [8].\\
Update the shortest path from source node to node [8] with distance 1.\\
The neighboring nodes of the node 8: [6, 7].\\
Update the shortest path from source node to node [6] with distance 2.\\
The neighboring nodes of the node 6: [3, 4, 8].\\
Update the shortest path from source node to node [3, 4] with distance 3.\\
The neighboring nodes of the node 3: [1, 2, 4, 6].\\
Update the shortest path from source node to node [1, 2] with distance 4.\\
The neighboring nodes of the node 1: [0, 3].\\
Update the shortest path from source node to node [0] with distance 5.\\
The neighboring nodes of the node 0: [1, 9].\\
Update the shortest path from source node to node [9] with distance 6.\\
The neighboring nodes of the node 9: [0, 5].\\
Update the shortest path from source node to node [5] with distance 7.\\
The farthest target from the source 7 is node 5 with distance 7.\\
The diameter path is [7, 8, 6, 3, 1, 0, 9, 5]. \\

\bottomrule

\label{table:Graph Diameter}
\end{tabular}

\vspace{-5mm}
\end{table*}

\begin{table*}[h]
\centering
\small
\caption{An example of the Maximum Clique Problem task.} 
\begin{tabular}{p{0.95\linewidth}}
\toprule

\textbf{Task Description}

In an undirected graph, (i,j) means that node i and node j are connected with an undirected edge. I'll give an instance of a graph, please help me find the maximum clique in the graph and list the steps and results for each iteration. The Maximum Clique Problem is an optimization problem in graph theory that aims to identify the largest set of vertices in a graph, where every two vertices in the set are adjacent. \\\\

\textbf{Input}

The graph has 9 nodes. The nodes are numbered from 0 to 8, and the edges are: [(2, 8), (3, 6), (4, 5), (5, 7)]. \\\\

\textbf{Output}

Add the most appropriate node: 8.\\
The current clique is: [8].\\
The common neighbors of nodes in the current clique are: [2].\\
Add the most appropriate node: 2.\\
The current clique is: [8, 2].\\
Finished!\\
The maximum clique is [8, 2]. \\

\bottomrule

\label{table:Maximum Clique Problem}
\end{tabular}

\vspace{-5mm}
\end{table*}

\begin{table*}[h]
\centering
\small
\caption{An example of the Traveling Salesman Problem task.} 
\begin{tabular}{p{0.95\linewidth}}
\toprule

\textbf{Task Description}

In an undirected graph, (i,j,k) means that node i and node j are connected with an undirected edge with weight k. I'll give an instance of a graph, please help me find the solution of the TSP problem in the given graph and list the steps and results for each iteration. The Traveling Salesman Problem (TSP) is a classic combinatorial optimization problem where, given a set of cities(nodes), the goal is to find the shortest possible route that visits each city(node) exactly once and returns to the starting city(node). For each iteration, please select the most appropriate node considering it's distance from the previous node and it's influence of total travel distance. \\\\

\textbf{Input}

The graph has 7 nodes. The nodes are numbered from 0 to 6, and the edges are: [(0, 1, 8309), (0, 2, 3986), (0, 3, 2254), (0, 4, 1983), (0, 5, 396), (0, 6, 2655), (1, 2, 1416), (1, 3, 9346), (1, 4, 3061), (1, 5, 3220), (1, 6, 7309), (2, 3, 8945), (2, 4, 6117), (2, 5, 9132), (2, 6, 4310), (3, 4, 7830), (3, 5, 1095), (3, 6, 3040), (4, 5, 9538), (4, 6, 6771), (5, 6, 1899)]. \\\\

\textbf{Output}

Choose starting node: 0.\\
Choose node 4 after node 0 with weight 1983.\\
The current subtour is [0, 4].\\
Choose node 1 after node 4 with weight 8309.\\
The current subtour is [0, 4, 1].\\
Choose node 2 after node 1 with weight 6117.\\
The current subtour is [0, 4, 1, 2].\\
Choose node 6 after node 2 with weight 7309.\\
The current subtour is [0, 4, 1, 2, 6].\\
Choose node 3 after node 6 with weight 8945.\\
The current subtour is [0, 4, 1, 2, 6, 3].\\
Choose node 5 after node 3 with weight 1899.\\
The current subtour is [0, 4, 1, 2, 6, 3, 5].\\
Choose node 0 after node 5 with weight 2254.\\
The current subtour is [0, 4, 1, 2, 6, 3, 5, 0].\\
Finished!\\
The optimal solution of TSP is: [0, 4, 1, 2, 6, 3, 5, 0]. \\

\bottomrule

\label{table:Traveling Salesman Problem}
\end{tabular}

\vspace{-5mm}
\end{table*}

\begin{table*}[h]
\centering
\small
\caption{An example of the Maximum Common Subgraph task.} 
\begin{tabular}{p{0.95\linewidth}}
\toprule

\textbf{Task Description}

In an undirected graph, (i,j) means that node i and node j are connected with an undirected edge, we call i and j are neighbors. Please find the maximum common subgraph of two graphs and list the steps and results for each iteration. The maximum common subgraph problem is defined as follows: given two graphs G and H, the task is to find a maximum subgraph sub\_g of G and a subgraph sub\_h of H such that sub\_g and sub\_h are isomorphic. That means some nodes of G and H should be put in the nodes lists sub\_g\_nodes and sub\_h\_nodes respectively satisfying the following 3 simple requirements. Firstly, sub\_g\_nodes and sub\_h\_nodes has the equal length. Secondly, any two nodes in sub\_g\_nodes or sub\_h\_nodes are different. Thirdly, for any available indices i and j, if (sub\_g\_nodes[i], sub\_g\_nodes[j]) is an edge of G, then (sub\_h\_nodes[i], sub\_h\_nodes[j]) must be also an edge of H, and vice versa. The objective is to maximize the length of sub\_g\_nodes. For each iteration, please choose a node u of G and another node v of H such that the indices of u'neighbors in sub\_g\_nodes must be same to the indices of v's neighbors in sub\_h\_nodes. \\\\

\textbf{Input}

The graph G has 5 nodes, the nodes are numbered from 0 to 4, and the edges are: [(0, 4), (1, 2), (1, 3), (1, 4), (3, 4)]. The graph H has 5 nodes, the nodes are numbered from 0 to 4, and the edges are: [(0, 1), (0, 2), (0, 3), (1, 2), (1, 3), (1, 4), (2, 4)]. \\\\

\textbf{Output}

Choose node 0 of G, and node 3 of H that has a similar neighborhood structure.\\
The current nodes lists of subgraphs are: [0], [3].\\
In sub\_g\_nodes, node 1 does not connect any node.\\
In sub\_h\_nodes, node 2 does not connect any node.\\
So choose node 1 of G, and node 2 of H as indices of their individual neighbors in the corresponding nodes lists are the same.\\
The current nodes lists of subgraphs are: [0, 1], [3, 2].\\
In sub\_g\_nodes, node 3 connects nodes of indices [1] which are [1] in G, and does not connect nodes of indices [0] which are [0] in G.\\
In sub\_h\_nodes, node 4 connects nodes of indices [1] which are [2] in H, and does not connect nodes of indices [0] which are [3] in H.\\
So choose node 3 of G, and node 4 of H as indices of their individual neighbors in the corresponding nodes lists are the same.\\
The current nodes lists of subgraphs are: [0, 1, 3], [3, 2, 4].\\
In sub\_g\_nodes, node 4 connects all nodes which are [0, 1, 3] in G.\\
In sub\_h\_nodes, node 1 connects all nodes which are [2, 3, 4] in H.\\
So choose node 4 of G, and node 1 of H as indices of their individual neighbors in the corresponding nodes lists are the same.\\
The current nodes lists of subgraphs are: [0, 1, 3, 4], [3, 2, 4, 1].\\
Finished!\\
The optimal solution of MCS is: [0, 1, 3, 4], [3, 2, 4, 1]. \\

\bottomrule

\label{table:Maximum Common Subgraph}
\end{tabular}

\vspace{-5mm}
\end{table*}

\begin{table*}[h]
\centering
\small
\caption{An example of the Graph Edit Distance task.} 
\begin{tabular}{p{0.95\linewidth}}
\toprule

\textbf{Task Description}

In an undirected graph, (i,j) means that node i and node j are connected with an undirected edge, we call i and j are neighbors. Next, I'll give you two special graphs, where each node has a label. You are required to solve the Graph Edit Distance problem between two graphs. The problem is to establish a one-to-one mapping between nodes from graph G to graph H, ensuring that each node in graph G corresponds to exactly one node in graph H, vice versa. For convenience, the notation i~j denotes one node mapping from node i in graph G to node j in graph H, also abbreviated as map(i)=j. For the node mapping i~j, if labels of node i and node j are the same, then the cost associated with i~j is 0, otherwise the cost is 1. For any two node mapping i~j and p~q from graph G to graph H, if (i,p) is an edge of G and (j,q) is not an edge of H, then i~j and p~q incur an edge deletion cost of 1, else if (i,p) is not an edge of G and (j,q) is an edge of H, then the edge insertion cost is 1, otherwise the edge cost is 0. The total cost is the sum of all node mapping costs and all edge costs. The objective is to minimize the total cost of a node mapping from G to H. You should represent the node mapping with an integer list L, such that node i~L[i] for all i.
 For instance, if node 0 in graph G corresponds to node 1 in graph H, node 1 in graph G corresponds to node 0 in graph H, and node 2 remains unchanged, the mapping would be represented as [1, 0, 2, ...]. \\\\

\textbf{Input}

The graph G has 5 nodes, the nodes are: [(0, 'Si'), (1, 'O'), (2, 'Si'), (3, 'Na'), (4, 'S')], and the edges are: [(0, 2), (0, 4), (1, 2), (1, 3), (1, 4), (2, 4), (3, 4)]. The graph H has 5 nodes, the nodes are: [(0, 'C'), (1, 'Na'), (2, 'N'), (3, 'Mg'), (4, 'N')], and the edges are: [(0, 1), (0, 2), (0, 3), (1, 3), (1, 4), (2, 3), (2, 4)]. \\\\

\textbf{Output}

Mapping node 0 labeled \textless Si\textgreater  of graph G to node 4 labeled \textless N\textgreater  of graph H.\\
As the mapping two nodes 0~4 have different labels, the node mapping cost adds 1.\\
The current mapping is [4] with cost 1.\\
Mapping node 1 labeled \textless O\textgreater  of graph G to node 0 labeled \textless C\textgreater  of graph H.\\
As the mapping two nodes 1~0 have different labels, the node mapping cost adds 1.\\
The current mapping is [4, 0] with cost 2.\\
Mapping node 2 labeled \textless Si\textgreater  of graph G to node 2 labeled \textless N\textgreater  of graph H.\\
As the mapping two nodes 2~2 have different labels, the node mapping cost adds 1.\\
The current mapping is [4, 0, 2] with cost 3.\\
Mapping node 3 labeled \textless Na\textgreater  of graph G to node 1 labeled \textless Na\textgreater  of graph H.\\
As the mapping two nodes 3~1 have the same label, the node mapping cost adds 0.\\
Currently for any index u of [0], node u does not connect node 3 in graph G, but map(u)=L[u] connects to map(3)=1 in graph H, so the new node mapping 3~1 generate edge addition cost 1.\\
The current mapping is [4, 0, 2, 1] with cost 4.\\
Mapping node 4 labeled \textless S\textgreater  of graph G to node 3 labeled \textless Mg\textgreater  of graph H.\\
As the mapping two nodes 4~3 have different labels, the node mapping cost adds 1.\\
Currently for any index u in [0], node u connects node 4 in graph G, but map(u)=L[u] does not connect to map(4)=3 in graph H, so the new node mapping 4~3 generate edge deletion cost 1.\\
The current mapping is [4, 0, 2, 1, 3] with cost 6.\\
Finished!\\
The optimal mapping of GED is: [4, 0, 2, 1, 3]. \\

\bottomrule

\label{table:Graph Edit Distance}
\end{tabular}

\vspace{-5mm}
\end{table*}